\documentclass{article}
\usepackage{arxiv}

\usepackage[utf8]{inputenc}
\usepackage[T1]{fontenc} 
\usepackage{pifont,amsmath,amssymb,amsfonts,nicefrac,microtype,lipsum,graphicx,natbib,booktabs,hyperref,url,doi}
\usepackage{cleveref}
\usepackage{authblk}
\usepackage{multibib}
\newcites{supp}{Supplementary References}
\usepackage[dvipsnames, svgnames, x11names]{xcolor}

\usepackage{xspace}

\newcommand{\red}[1]{{\color{red}#1}}

\definecolor{caribbeangreen}{rgb}{0.0, 0.8, 0.6}
\definecolor{coralred}{rgb}{1.0, 0.25, 0.25}
\definecolor{royalblue(web)}{rgb}{0.25, 0.41, 0.88}
\definecolor{jade}{rgb}{0.0, 0.66, 0.42}
\definecolor{jasper}{rgb}{0.84, 0.23, 0.24}
\definecolor{amber}{rgb}{1.0, 0.75, 0.0}

\newcommand{\encoder}{\ensuremath{\mathcal{E}}\xspace}
\newcommand{\encodersoft}{\ensuremath{\tilde{\mathcal{E}}}\xspace}
\newcommand{\decoder}{\ensuremath{\mathcal{P}}\xspace}
\newcommand{\generator}{\ensuremath{\mathcal{G}}\xspace}

\newcommand{\decodersub}{\ensuremath{\mathcal{P}_{\text{sub}}}\xspace}
\newcommand{\inimage}{\ensuremath{I}\xspace} 
\newcommand{\inimagepred}{\ensuremath{\hat{I}}\xspace} 

\newcommand{\gaze}{\ensuremath{\hat{g}}\xspace}
\newcommand{\valid}{\ensuremath{\hat{v}}\xspace}

\newcommand{\validgt}{\ensuremath{v}\xspace}
\newcommand{\gazegt}{\ensuremath{g}\xspace}

\newcommand{\teacher}{\mathcal{P}_{\mathrm{T}}}
\newcommand{\controller}{C_{\theta}}

\newcommand{\hstudent}{\mathbf{h}}
\newcommand{\hteacher}{\mathbf{h}_{\mathrm{T}}}

\newcommand{\compactlatent}{\mathbf{h}}
\newcommand{\compactlatentedit}{\Delta \mathbf{h}}
\newcommand{\fullwplus}{\mathbf{w}^{+}}
\newcommand{\fullwplusedit}{\Delta \mathbf{w}^{+}}

\newif\ifarxiv
\arxivtrue 
\newcommand{\Description}[2][]{}

\begin{document}

\title{Low Latency Gaze Tracking via Latent Optical Sensing}

\renewcommand\Authfont{\bfseries}
\setlength{\affilsep}{0em}

\author{Yidan Zheng$^*$}
\author{Matheus Souza$^*$}
\author{Kaizhang Kang}
\author{Qiang Fu}
\author{Hadi Amata}
\author{Wolfgang Heidrich}

\affil{King Abdullah University of Science and Technology (KAUST)}
\date{}

\maketitle
\def\thefootnote{*}\footnotetext{These authors contributed equally to this work}

\begin{figure*}[h]
\centering
  \includegraphics[width=1\textwidth]{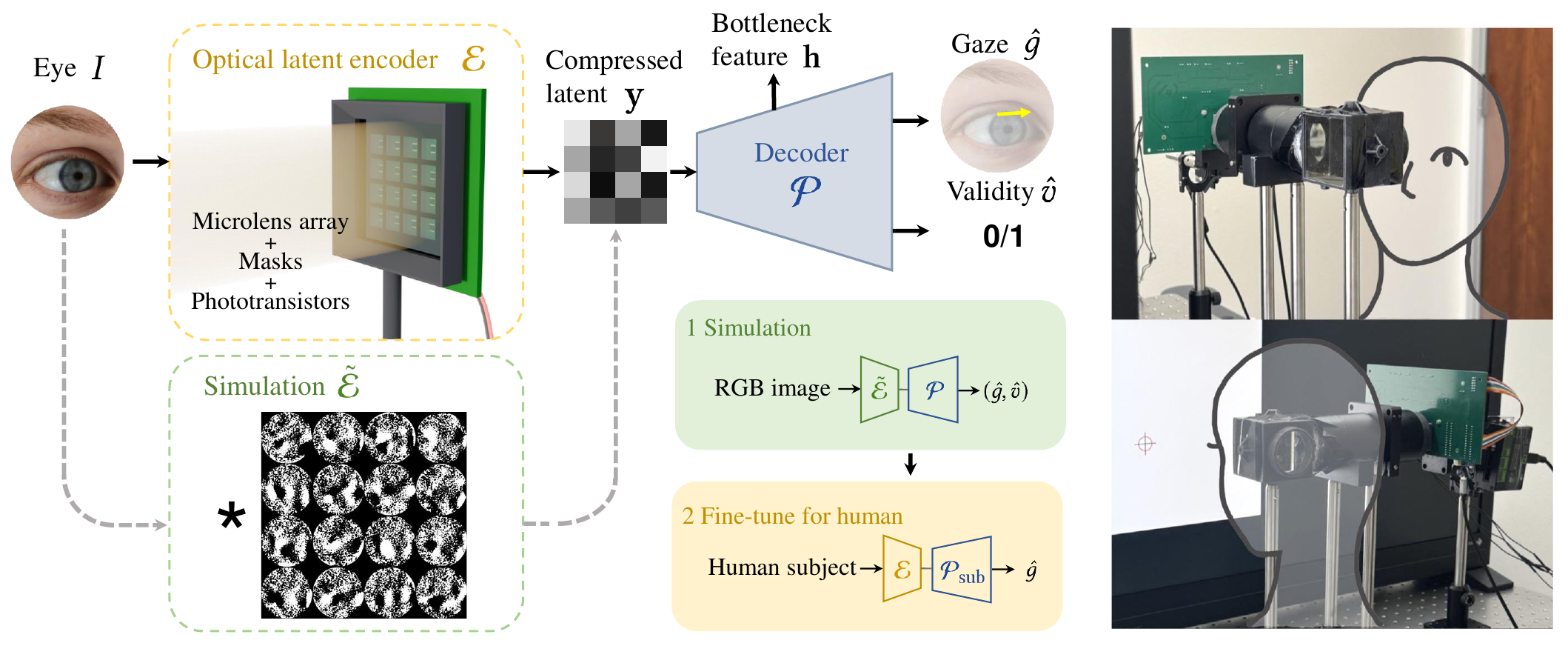}
  \caption{\textbf{Overview of low-latency gaze tracking system.} Our system replaces conventional high-resolution cameras with a fully passive optical latent encoder $\mathcal{E}$. Light from the eye $I$ is modulated by a microlens array and co-designed binary masks, producing a compressed latent measurement $c$ captured by a 16 element phototransistor array. These features are mapped to gaze direction $\hat{g}$ and validity $\hat{v}$ via a lightweight MLP $\mathcal{F}$ and predictor $\mathcal{P}$. Our end-to-end pipeline (bottom center) uses simulation-to-real transfer with human-subject fine-tuning to achieve a total latency of less than $4~\mathrm{ms}$. (Right) Experimental setup and hardware implementation.}
  \Description[Overview of low-latency gaze tracking system]
{A compact optical gaze tracking system that uses a microlens array and patterned masks to encode eye images into a small set of intensity measurements. These measurements are directly mapped to gaze direction using a lightweight neural network, enabling real-time inference with very low latency. The figure illustrates the full pipeline from eye input, optical encoding, sensor measurements, to gaze output.}
  \label{fig:teaser}
\end{figure*}

\begin{abstract}

We present a real-time gaze tracking system that directly acquires task-relevant latent features using a fully passive optical encoder. Instead of forming and processing full-resolution images, our approach leverages a microlens array with a co-designed binary chromium mask to perform spatially multiplexed optical encoding, producing a compact set of measurements sufficient for gaze estimation. By integrating sensing and feature extraction in the optical domain, the proposed system eliminates the need for high-bandwidth image readout and substantially reduces computational overhead. The encoded measurements are captured by a $4 \times 4$ phototransistor array and mapped to gaze direction using a lightweight neural network. Our proof-of-concept prototype enables an end-to-end sensing-to-inference latency of  $3.4~\mathrm{ms}$, outperforming published research systems. We demonstrate the effectiveness of our approach on both simulated and real-world data, achieving competitive gaze estimation accuracy while significantly reducing end-to-end latency compared to conventional camera-based pipelines. The proposed system establishes a new operating point in the accuracy–latency–compute trade-off for latency- and resource-constrained gaze tracking. This work highlights the potential of task-driven optical sensing for ultra-low-latency, computationally efficient human-computer interaction systems.

\end{abstract}


\section{Introduction}\label{sec:introduction}

Eye gaze tracking plays an important role in applications such as human--computer interaction~\cite{smith2013gaze,zhang15_cvpr,kellnhofer2019gaze360}, assistive technologies~\cite{10.1145/507072.507076,caligari2013eye}, and emerging AR/VR systems~\cite{tanriverdi2000interacting,ugwitz2022eye,shadiev2023review, stengel2015affordable}. In particular, next-generation wearable displays require gaze estimation systems that are compact, energy-efficient, and capable of operating at ultra-low latency to support real-time interaction and foveated rendering~\cite{10.1117/12.2322657, 10.1145/2366145.2366183}. Conventional appearance-based gaze tracking systems typically rely on camera-based image acquisition followed by deep neural network inference~\cite{kellnhofer2019gaze360,davalos2025webeyetrack,muksimova2025gazecapsnet}. While recent advances in deep learning have significantly improved gaze estimation accuracy~\cite{Zhang2020ETHXGaze, barkevich2024using, zhang19_pami, abdelrahman2023l2cs}, these approaches still require capturing and processing full-resolution eye images, introducing substantial sensing, bandwidth, and computational overhead.

To address these limitations, recent research has explored lightweight and event-driven gaze sensing systems. Event-based approaches leverage asynchronous vision sensors to reduce redundant image acquisition and improve temporal resolution~\cite{li2024egaze,Bonazzi_2024_CVPR,9389490,10.1145/3699745,10.5555/3666122.3668838,10918853}. Other works investigate embedded and edge-based gaze tracking pipelines for low-power operation~\cite{10325167}. However, most existing systems still depend on reconstructing or processing image-like measurements before gaze inference, resulting in additional latency and computational cost. Furthermore, event-based methods often define latency only as the processing time after event accumulation~\cite{li2024egaze}, without accounting for the sensing time required to collect sufficient events under varying illumination conditions.

At the same time, advances in computational imaging and task-driven optical design suggest that optical systems can be optimized directly for downstream inference objectives rather than image quality~\cite{shi2022loen, minimalist_camera, souza2025latentspaceimaging}. Instead of treating images as mandatory intermediate representations, these approaches aim to capture compact task-relevant measurements that preserve the information necessary for a specific vision task. This perspective is particularly attractive for gaze tracking, where the final target is a low-dimensional gaze vector rather than a visually interpretable eye image. Rather than replacing high-accuracy camera-based gaze trackers, our goal is to explore a different operating point that prioritizes minimal sensing complexity and ultra-low end-to-end latency, targeting applications such as wearable devices, always-on gaze sensing, coarse gaze interaction, and attention estimation.

Motivated by these observations, we propose a passive optical gaze tracking system that directly acquires low-dimensional latent measurements using a microlens array (MLA) and optimized binary masks. Unlike conventional camera-based gaze trackers, the proposed system avoids explicit image formation and instead performs task-oriented optical encoding in a single shot. The encoded optical responses are measured by phototransistors and mapped to gaze direction using a lightweight neural network. By eliminating high-resolution image acquisition and minimizing computational overhead, the proposed system achieves ultra-low-latency gaze estimation with a compact and passive optical architecture.


To enable robust latent-space sensing, we develop an end-to-end training framework that jointly optimizes the optical encoding masks and digital inference using both latent reconstruction and gaze supervision objectives. The learned optical encoder is trained to preserve gaze-relevant information within a compact 16-dimensional measurement space while remaining compatible with passive optical implementation. We further evaluate the system on both synthetic and real human eye data and analyze its end-to-end latency under realistic hardware conditions.

The main contributions of this work are summarized as follows:
\begin{itemize}
\item We propose an end-to-end latent-space gaze tracking framework that jointly optimizes passive optical encoding and neural gaze decoding, enabling direct gaze inference from compact low-dimensional measurements without reconstructing full eye images.

\item We develop a compact MLA-based single-shot optical sensing system with optimized binary masks, phototransistor readout, and field-programmable gate array (FPGA) acquisition, enabling low sensing complexity and an end-to-end latency of $3.4\,\mathrm{ms}$ including both sensing and computation.

\item We establish a two-stage simulation-to-real evaluation \\pipeline, achieving a cross-subject simulation error of $6.47^\circ$ and a calibrated per-subject real-world error below $6^\circ$.
\end{itemize}

Although our proof-of-concept prototype is not engineered to optimize
for overall system power consumption, we note that the substantial
reduction of sensor measurements, the resulting reduction in data
volume, and the extremely lightweight nature of the digital processing
should allow for highly energy efficient implementations of the
concept in the future.

\section{Related Work}\label{sec:related_works}

\paragraph{Low-latency and real-time gaze estimation.}
Recent research has underscored that for mobile XR glasses, achieving
an end-to-end latency of less than $5~\mathrm{ms}$ is a critical
threshold to prevent the disruption of user
immersion~\cite{10.1117/12.2322657}. While many software-oriented
frameworks such as \textit{WebEyeTrack}~\cite{davalos2025webeyetrack}
and \textit{GazeCapsNet}~\cite{muksimova2025gazecapsnet} report
exceptionally fast inference speeds via few-shot personalization or
lightweight neural architectures, they typically rely on full-image
capture from conventional CMOS sensors. This reliance introduces
significant power and temporal overheads due to frame-rate limitations
and large-scale data readout, making them less efficient for truly
real-time applications. To address these bottlenecks,
hardware-oriented approaches have explored alternative sensing
modalities. Li et al.~\cite{9284794} demonstrated that spatially
sparse single-pixel detectors can substitute for dense imaging
sensors, while visible light sensing was proposed for
energy-constrained headsets~\cite{10.1145/3131672.3131682}.On the
edge, systems like \textit{TinyTracker}~\cite{10325167} utilize
in-sensor processing to enable fast, low-power operation on mobile
platforms.

However, the system latency is limited not just by the computational
processing, but also the exposure and readout time for the full
frame. Moreover, 2D image sensor arrays are relatively power hungry
due to the large number of A/D conversion needed. Event-based
sensing~\cite{li2024egaze, 10.1145/3699745, Bonazzi_2024_CVPR} has
been proposed to increase the frame rate and reduce the latency.
However, event-based gaze estimation still relies on temporal event accumulation and subsequent digital processing before inference, and reported latencies are often measured after sufficient events have been acquired~\cite{10.5555/3666122.3668838,9389490}. Other work such as
Pixel Processor Arrays~\cite{9757572} and dedicated ASICs like
\textit{JaneEye}~\cite{janeeye2025} have focused primarily on the
sensing hardware, and achieve impressive sensing latency as low as
$0.5~\mathrm{ms}$. However, for a complete eye tracking system, these
sensors need to be paired heavy compute architectures requiring tens
of MFLOPs per frame. In contrast, our system directly produces task-specific optical measurements rather than image frames or event streams, reducing sensing bandwidth and digital processing while achieving an end-to-end latency below $4~\mathrm{ms}$ with only $0.49~\mathrm{MFLOPs}$.

At the algorithmic level, several works have pointed out that full eye images are highly redundant for gaze estimation. For example, EyeTrAES~\cite{10.1145/3699745} introduces adaptive event slicing to extract only relevant temporal features, while dataset-driven and representation learning approaches~\cite{10.1145/3290605.3300780, sun2021cross, lee2022latentgaze, Wang_2022_CVPR, ghosh2022mtgls, cheng2022puregaze} focus on condensing eye features into compact, generalizable codes. These methods consistently emphasize that raw image data is unnecessary overhead for the task. Our work follows this trajectory by eliminating full image formation entirely, instead acquiring compact latent measurements directly through task-driven optical encoding.

\paragraph{Task-driven optics and latent-space sensing.}
Recent advances in computational imaging have demonstrated the effectiveness of jointly optimizing optical systems and downstream algorithms. By incorporating differentiable models of wave propagation, these approaches enable end-to-end learning of optical elements together with reconstruction networks, often targeting high-fidelity image formation~\cite{ho2024differentiablewaveoptics, dai2025toleranceawaredeepoptics}. This paradigm has been explored in a variety of settings, including learned diffractive optics~\cite{wang2023image}, task-driven lens design~\cite{yang2026}, collaborative camera arrays~\cite{sun2025collaborative}, task-specific encoding for depth and RGBD sensing~\cite{liu2025learned}, and coded aperture~\cite{salman2017flatcam, zhao2023coded} or compressive imaging systems~\cite{duarte2008single}, where measurements are optimized for efficient reconstruction.

Beyond image formation, a growing body of work shifts the objective from reconstruction to task performance. In these systems, optical front-ends are optimized directly for downstream objectives such as classification or detection, as demonstrated in lensless opto-electronic neural networks~\cite{shi2022loen}, freeform optical designs~\cite{minimalist_camera}, and compressive and single-pixel imaging systems~\cite{10.1007/978-3-031-72904-1_27}. This perspective recognizes that full image recovery is often unnecessary when the goal is to extract semantic information.

More recently, latent-space sensing extends this idea by learning optical encoders that map inputs directly into compact feature representations, which are then processed by lightweight neural networks~\cite{souza2025latentspaceimaging}. By bypassing explicit image reconstruction, such approaches improve efficiency and reduce redundancy in the sensing pipeline.

Building on this perspective, we explore latent-space sensing for gaze tracking, where the optical front-end directly encodes low-dimensional gaze features instead of forming eye images. This formulation enables a compact and efficient sensing pipeline tailored for real-time applications such as AR/VR.

\section{Gaze Tracking Framework and Dataset}\label{sec:model}

Conventional image-based gaze estimation systems typically rely on densely sampled eye images, requiring the sensor to capture, digitize, and process a large number of pixels. 
However, the information needed for gaze estimation may lie on a much lower-dimensional manifold than the full image representation. 
Motivated by this observation, we estimate gaze from only $16$ scalar measurements acquired in a single snapshot.

As shown in Figure~\ref{fig:teaser}, our pipeline learns optical sensing patterns that are optimized directly for gaze estimation. The optimized optical encoder $\encoder$ acquires a measurement vector $\mathbf{y} \in \mathbb{R}^{16}$. 
This measurement is then processed by a lightweight decoder $\decoder$, which is trained to predict the latent representation $\compactlatent$ of a pretrained generative model $\generator$, together with gaze $\gaze$ and a validity flag $\valid$. 
Rather than targeting pixel-level reconstruction, the system is optimized to recover a compact latent representation that preserves gaze-relevant information.

\subsection{Optical Encoder and Latent Distillation}
\label{sec:optical_latent_encoder}

To train this system, we define a differentiable digital encoder $\encodersoft$ that emulates the measurement process of the optical encoder $\encoder$. 
Given an input eye image $\inimage$, $\encodersoft$ is implemented as a bank of $16$ learnable binary masks at the image resolution:
\begin{equation}
    \encodersoft = \{ \mathbf{m}_k \}_{k=1}^{16},
    \qquad
    \mathbf{m}_k \in \{0,1\}^{256 \times 256}.
\end{equation}
Each mask serves as an optical sensing pattern, and the corresponding scalar measurement is computed by an inner product:
\begin{equation}\label{eq:sum}
    y_k = \langle \inimage, \mathbf{m}_k \rangle,
    \qquad
    k = 1,\ldots,16.
\end{equation}
The full measurement vector is
\begin{equation}
    \mathbf{y}
    =
    [y_1, y_2, \ldots, y_{16}]^\top
    \in \mathbb{R}^{16}.
\end{equation}

Since the optical encoder provides only 16 scalar measurements, direct supervision from gaze labels alone may not be sufficient to learn a robust representation. 
Following~\cite{souza2025latentspaceimaging}, we train a student decoder $\decoder$ to recover a compact feature aligned with the latent space of a pretrained generative inversion model. 
Thus, the decoder learns not only from gaze labels, but also from a structured latent representation that captures gaze-relevant eye appearance.

To obtain the generative latent space used for supervision, we train a StyleGAN2 generator $\generator$ on eye patches~\cite{karras2020analyzing}, together with an inversion model $\teacher$ based on the encoder architecture of~\cite{richardson2021encoding}. 
The inversion model maps an eye image to a compact latent feature and serves as the teacher for $\decoder$ during distillation training process. 
Given a clean eye image $\inimage$, the teacher predicts a compact latent feature, gaze estimate and a validity logit:

\begin{equation}
    \hteacher, \gaze_T, \valid_{T} = \teacher(\inimage),
    \qquad 
    \hteacher \in \mathbb{R}^{256}.
\end{equation}

The student decoder $\decoder$ maps the compressed measurement vector $\mathbf{y}$ to a 256-dimensional bottleneck feature $\hstudent$ and then predicts both gaze and validity:
\begin{equation}
    \hstudent, \gaze, \valid = \decoder(\mathbf{y}),
    \qquad 
    \hstudent \in \mathbb{R}^{256}.
\end{equation}

The decoder consists of a normalization layer, an upscaling MLP block, and a two-layer MLP residual bottleneck, followed by separate gaze and validity heads; architectural details are provided in the supplementary material. 
The gaze and validity heads are initialized from the corresponding pretrained heads of $\teacher$ to accelerate convergence.

Although the student is not trained with an image reconstruction objective, its bottleneck feature remains connected to the generator latent space. 
In particular, the compact feature can be repeated across the $L=14$ StyleGAN layers to form a broadcast $\mathcal{W}^{+}$ code when synthesis or latent editing is required.

The training objective combines supervised gaze regression, validity classification, gaze distillation, and feature distillation. 
We denote the student-predicted gaze vector by $\gaze$, the ground-truth gaze by $\gaze_{\mathrm{GT}}$, and the teacher-predicted gaze by $\gaze_{T}$. 
Gaze errors are measured using cosine distance, i.e., one minus the cosine similarity between two 3D gaze vectors. 
The total loss is
\begin{equation}
\begin{split}
\mathcal{L} =
&\lambda_{\text{sup}}
\validgt \,
\mathcal{L}_{\text{cos}}(\gaze, \gaze_{\mathrm{GT}})
+
\lambda_{\text{cls}}
\text{BCE}(\valid, \validgt)
\\
&+
\lambda_{\text{gaze-distill}}
\validgt \,
\mathcal{L}_{\text{cos}}(\gaze, \gaze_{T})
+
\lambda_{\text{feat-distill}}
\validgt \,
\|\hstudent - \hteacher\|_2^2 .
\end{split}
\end{equation}
Here, the first two terms train the student using ground-truth gaze and validity labels, while the last two terms distill the teacher gaze prediction $\gaze_{T}$ and compact latent feature $\hteacher$. 
The gaze and feature terms are evaluated only on valid samples using $\validgt$, whereas the validity classification loss is applied to all samples.

This objective trains the student to recover a compact, gaze-aware latent representation directly from the 16 optical measurements, without requiring image reconstruction during this stage. 
Because this representation is aligned with the teacher latent feature and remains compatible with the generator through the broadcast $\mathcal{W}^{+}$ construction, it also provides a natural latent coordinate system for gaze control.

\subsection{Gaze Steering in Latent Space}
\label{sec:gaze_steering}

To evaluate the ability of the latent space to effectively encode gaze directions and separate the corresponding parameters from appearance and the visual identity of the subject, we demonstrate gaze steering using the trained generator and residual displacement in latent space (see supplemental videos for examples). This experiment is intended as an analysis of the learned latent representation rather than part of the gaze estimation pipeline. Successful gaze steering indicates that the recovered latent preserves gaze-related semantic information.

Given a source gaze $\gaze_s$ and a target gaze $\gaze_t$, a control network $\controller$ predicts a latent update that moves the representation from the source gaze direction toward the target direction.

For the compact latent representation used above, the controller predicts
\begin{equation}
    \compactlatentedit = \controller(\gaze_s,\gaze_t),
    \qquad
    \compactlatentedit \in \mathbb{R}^{256}.
\end{equation}
The edited compact latent code is then
\begin{equation}
    \compactlatent' = \compactlatent + \compactlatentedit,
\end{equation}
where $\compactlatent$ is the compact latent representation produced by the inversion encoder. 
The edited latent can be decoded by the frozen generator after being repeated across the StyleGAN layers using the same broadcast construction described above.

For higher-fidelity gaze steering, we also consider editing directly in the full StyleGAN $\mathcal{W}^{+}$ space of the color-image generator. 
In this setting, the latent representation consists of $L=14$ independent style vectors, each of dimension 512. 
The controller therefore predicts a layer-wise residual
$\fullwplusedit \in \mathbb{R}^{14 \times 512}$, which is added to the inverted latent code before synthesis:
\begin{equation}
    \inimage' = G(\fullwplus + \fullwplusedit).
\end{equation}
This full $\mathcal{W}^{+}$ formulation follows the same residual-control principle as the compact case, but provides additional degrees of freedom by allowing each StyleGAN layer to receive its own gaze-dependent update.

In practice, $\controller$ takes the concatenated gaze pair $[\gaze_s;\gaze_t]$ as input. 
A shared multilayer perceptron encodes this pair into a hidden representation, and the output is either a single compact displacement $\compactlatentedit$ or a set of layer-wise displacements $\fullwplusedit$. 
The generator and inversion encoder remain frozen throughout training; only the control network is optimized to produce latent edits that steer gaze while preserving identity and eye appearance.

\begin{figure*}[t]
  \centering
   \includegraphics[width=0.80\linewidth]{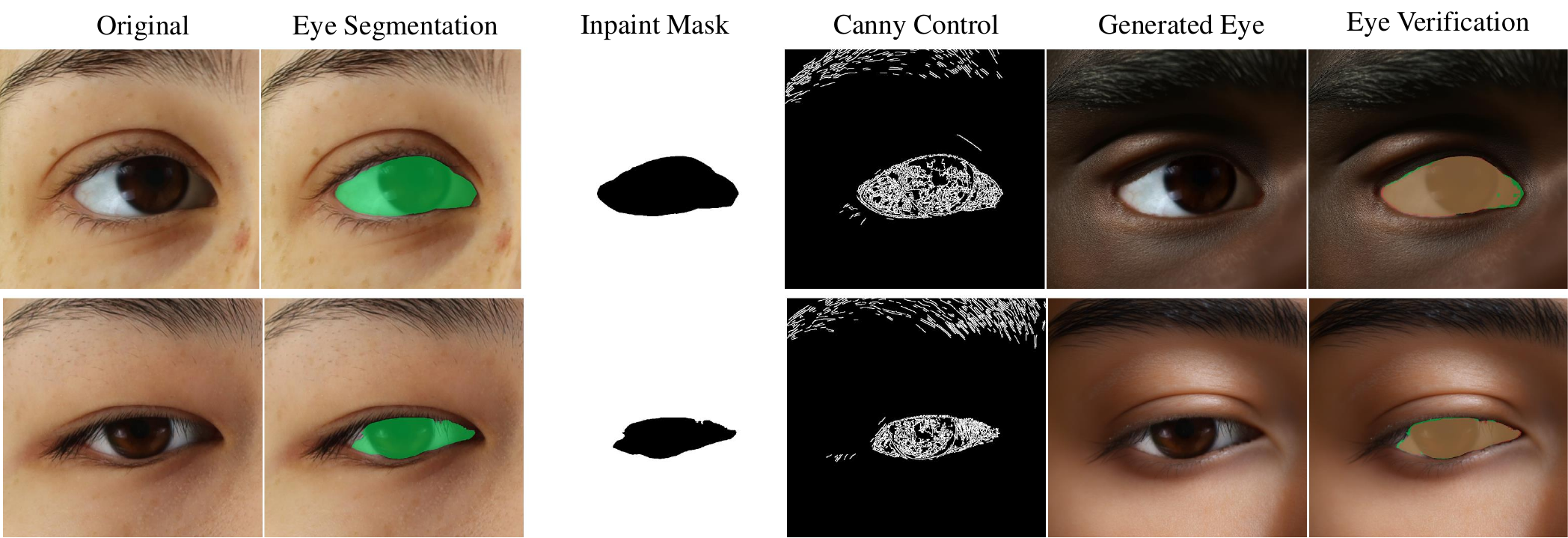}
\caption{\textbf{Identity-varying augmentation via generative
inpainting.} From left to right: original eye crop, SAM3 eye segmentation, inpainting mask, Canny-edge ControlNet input, the Flux-generated eye (conditioned on the mask, the Canny edges, and a diversity prompt), and SAM3 re-segmentation of the generated eye overlaid on the original mask for consistency checking. Samples failing the area/IoU check are regenerated with a different seed for up to three attempts; if none passes, the original crop is retained. The gaze label therefore always remains valid. More augmentation examples can be found in the Supplementary Material.}
  \Description[Identity-varying augmentation via generative
inpainting]
  {From left to right: original eye crop, SAM3 eye segmentation, inpainting mask, Canny-edge ControlNet input, Flux-generated eye conditioned on the mask, Canny edges, and a diversity prompt, and SAM3 re-segmentation of the generated eye overlaid on the original mask for consistency checking. SSamples failing the area/IoU check are regenerated with a different seed for up to three attempts; if none passes, the original crop is retained.}
   \label{fig:augmentation}
\end{figure*}

\begin{figure*}[t]
  \centering
  \includegraphics[width=0.8\linewidth]{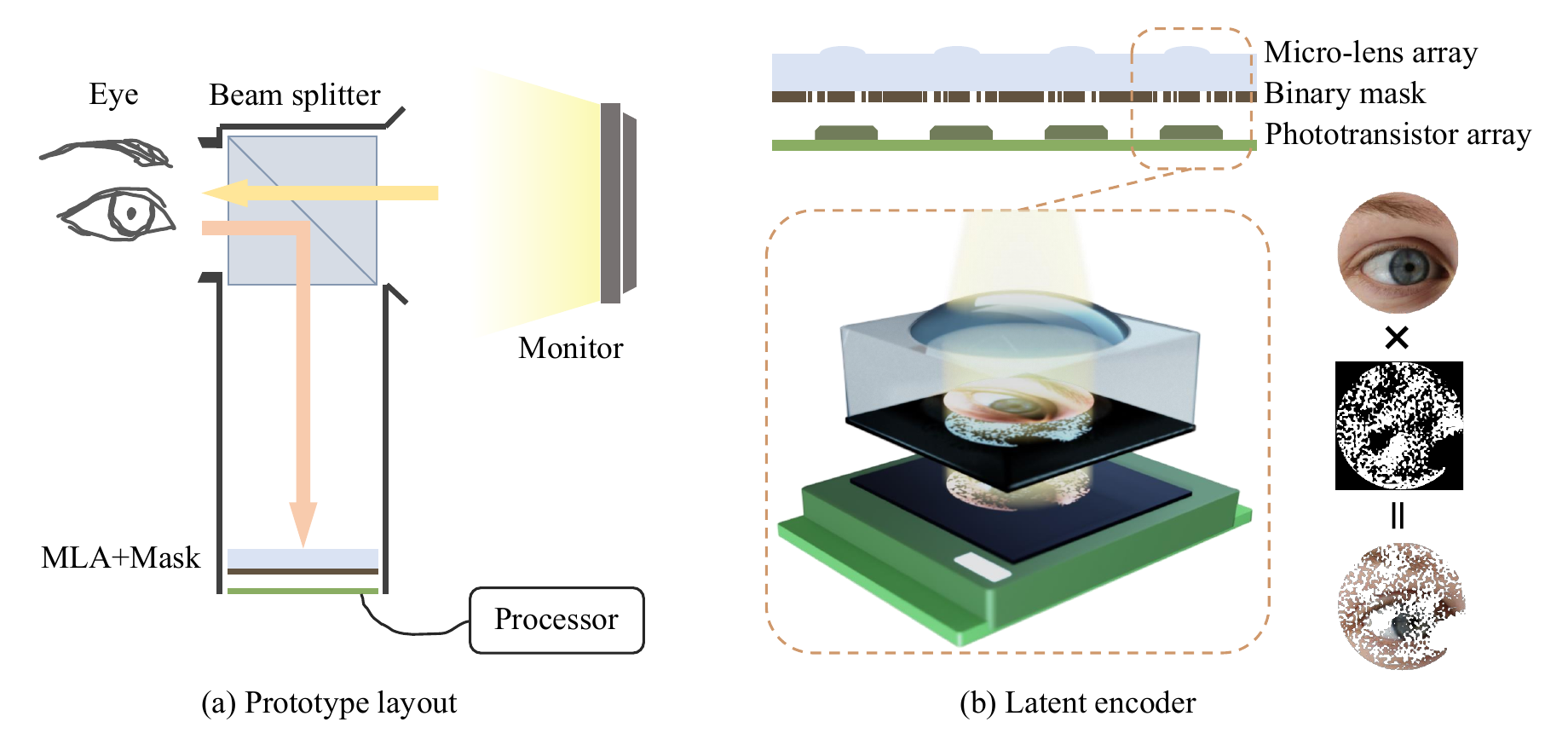}
  \caption{\textbf{Hardware prototype.} (a) The prototype layout utilizes a beam splitter to redirect the eye's reflection toward the MLA+Mask assembly while allowing the user to view a monitor. (b) Detailed view of the latent encoder stack, consisting of a micro-lens array, a binary chromium mask, and a phototransistor array. The encoder performs spatial multiplexing, effectively convolving the eye image with a task-specific binary pattern to produce compressed measurements for direct gaze estimation.}
  \Description{This image details a real-time gaze tracking prototype that performs feature acquisition entirely in the optical domain. The system employs a beam splitter to capture eye features through a custom latent encoder.}
  \label{fig:prototype}
\end{figure*}

\subsection{Dataset}
\label{sec:dataset}

Our dataset is derived from ETH-XGaze~\cite{Zhang2020ETHXGaze}, consisting of high-resolution monocular eye patches. We utilize a processing pipeline involving frontal camera selection and a circular masking process to mitigate microlens-induced corner artifacts. To ensure robustness against mechanical tolerances and optical alignment errors in the physical prototype, we apply random spatial translations, scaling, and horizontal flipping as data augmentation during training. Each sample is annotated with a 3D gaze vector and a binary validity label to identify blinks or misalignments. Further technical details on the cropping geometry, coordinate normalization, and the automated classification of invalid inputs are provided in the Supplementary Material. The final processed corpus includes 68 training subjects and 12 test subjects, totaling over 168,000 images. Over valid samples, gaze spans $-35.9^\circ$ to $22.5^\circ$ in pitch and $-54.7^\circ$ to $56.5^\circ$ in yaw. The held-out test split covers a slightly narrower range, $-31.7^\circ$ to $19.1^\circ$ in pitch and $-46.6^\circ$ to $47.2^\circ$ in yaw.

\subsection{Identity Augmentation via Generative Inpainting}
\label{sec:aug_gaze}
                                                     
Appearance-based gaze models are often prone to identity bias, learning shortcuts based on skin tone or facial structure rather than ocular geometry \cite{akgul2026gazeFairness}. This is exacerbated by the limited demographic diversity typical of laboratory-collected datasets \cite{Zhang2020ETHXGaze}. To address this, we introduce a generative augmentation scheme leveraging the Flux diffusion model \cite{flux2024} to diversify the periocular appearance associated with each ground-truth gaze label.

Our pipeline utilizes SAM3 \cite{carion2025sam3segmentconcepts} to segment the eye, creating an inpainting mask that preserves the original ocular pixels while allowing the surrounding context to be regenerated. To maintain anatomical plausibility, we employ Canny-edge \cite{4767851} ControlNet \cite{zhang2023adding} as a structural guide. Flux then inpaints the masked region using prompts spanning diverse ethnicities, ages, and lighting conditions. To ensure geometric fidelity, each sample must pass a CLIP-based \cite{radford2021learning} semantic check and an eye-consistency verification requiring 90\% spatial overlap ($\mathrm{IoU} \geq 0.9$) with the original segmentation. This process reduces the model's reliance on identity-correlated features, as illustrated in the pipeline overview in Figure~\ref{fig:augmentation}. Extended details on the prompt pool and verification metrics are provided in the Supplementary Material.

\section{Prototype}\label{sec:prototype}

\begin{figure*}[t]
  \centering
  \includegraphics[width=0.9\linewidth]{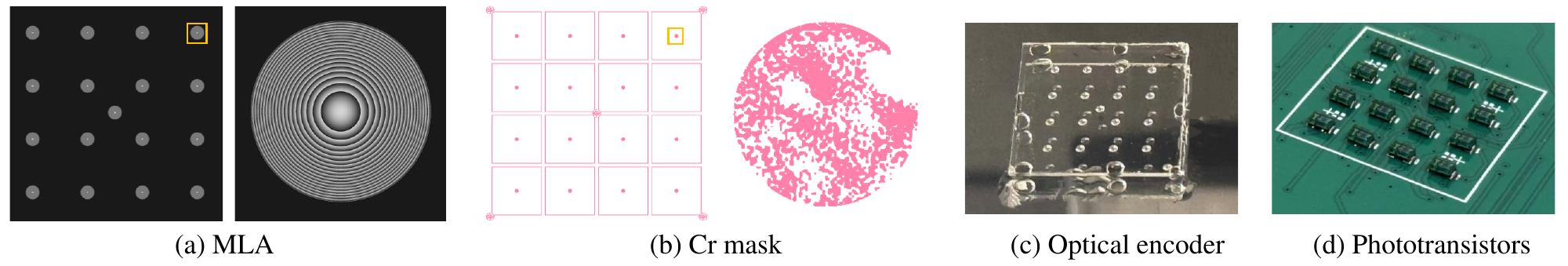}
  \caption{\textbf{Fabrication of the latent optical encoder.} (a) Design layout of the Fresnel microlens array (MLA) featuring a 1~mm aperture. (b) Patterned binary Cr mask designed for spatial modulation. (c) Photograph of the assembled latent encoder, where the MLA and Cr mask wafers are precisely aligned and bonded using UV-curable glue. (d) Photograph of the $4\times4$ phototransistor array integrated on a custom PCB for high-speed intensity acquisition.}
  \Description[Fabrication components of the latent optical encoder]
  {A four-part figure showing the hardware components. Part (a) is a grayscale top-down view of a Fresnel lens with concentric rings. Part (b) shows a circular pink and white stochastic binary pattern. Part (c) is a photograph of a transparent square substrate containing a grid of small optical elements. Part (d) shows a green printed circuit board with a four-by-four grid of small black surface-mount phototransistors.}
  \label{fig:fab}
\end{figure*}

Our prototype consists of a beam splitter and a hardware sensing module composed of a microlens array (MLA), a binary chromium (Cr) mask, and a phototransistor array, as illustrated in Fig.~\ref{fig:prototype}. The MLA and the Cr mask are fabricated on separate substrates and bonded together after precise alignment. In operation, the eye is illuminated and the reflected light is directed by the beam splitter into the Latent Encoder. The microlens array forms micro-images of the eye, which are selectively modulated by the binary mask. The resulting encoded intensity is captured by the phototransistor array, which is integrated with an FPGA-based acquisition circuit to yield ultra-low-latency latent optical measurements for gaze estimation.

\subsection{Microlens Array and Cr Mask}

\subsubsection{Design}
The sensing module utilizes an MLA composed of Fresnel lenses, each featuring an aperture of $1~\mathrm{mm}$ and a focal length of $1.94~\mathrm{mm}$. These lenses are designed to demagnify the circular input eye region (diameter of $40~\mathrm{mm}$) into a reduced spot of $0.256~\mathrm{mm}$, focused at the mask plane. The binary Cr mask is patterned with corresponding features of $0.256~\mathrm{mm}$ to match this demagnified image size. Our architecture employs a direct mapping where each phototransistor in the $4 \times 4$ array measures the integral of light intensity over the corresponding masked image as described in Eq.~\eqref{eq:sum}. The phototransistors are spaced with a $4~\mathrm{mm}$ horizontal pitch and a $3.88~\mathrm{mm}$ vertical pitch. The layout of MLA and Cr mask design is illustrated in Fig.~\ref{fig:fab}(a) and (b).

\subsubsection{Fabrication}
The MLA and the binary Cr mask were fabricated on separate wafers. The MLA ($6.332~\mathrm{\mu m}$ wrapping height, $1~\mathrm{\mu m}$ fabrication pitch) was fabricated on OrmoComp using grayscale direct-write lithography followed by UV nanoimprinting, following the methodology in \cite{Amata:24}.  The binary Cr mask ($1~\mathrm{\mu m}$ fabrication pitch) was patterned on a separate wafer using the same  Heidelberg DWL66+ direct-write lithography system. The MLA and mask substrates were then bonded together using careful optical alignment to place the mask near the MLA focal plane. The assembled prototype is shown in Fig.~\ref{fig:fab}(c).

\subsection{Phototransistors with FPGA system}
We employ an array of TEMT6200FX01 phototransistors, which feature a spectral sensitivity of 450--610~nm and a radiant sensitive area of $0.75~\mathrm{mm}^2$. To minimize electronic noise and ensure signal integrity, the supporting printed circuit board (PCB) includes an amplifier and a low-pass filter, followed by an analog-to-digital converter (ADC). The converted digital signals are read out by an FPGA at a stable sampling rate of 14,400~$\mathrm{Hz}$ ($900~\mathrm{Hz}$ for 16 phototransistors). This hardware configuration enables the rapid acquisition of latent features with high signal-to-noise ratio, supporting the system's low end-to-end latency.

\section{Evaluation}\label{sec:evaluation}

\begin{figure}[ht]
  \centering
    \ifarxiv
        \includegraphics[width=0.5\linewidth]{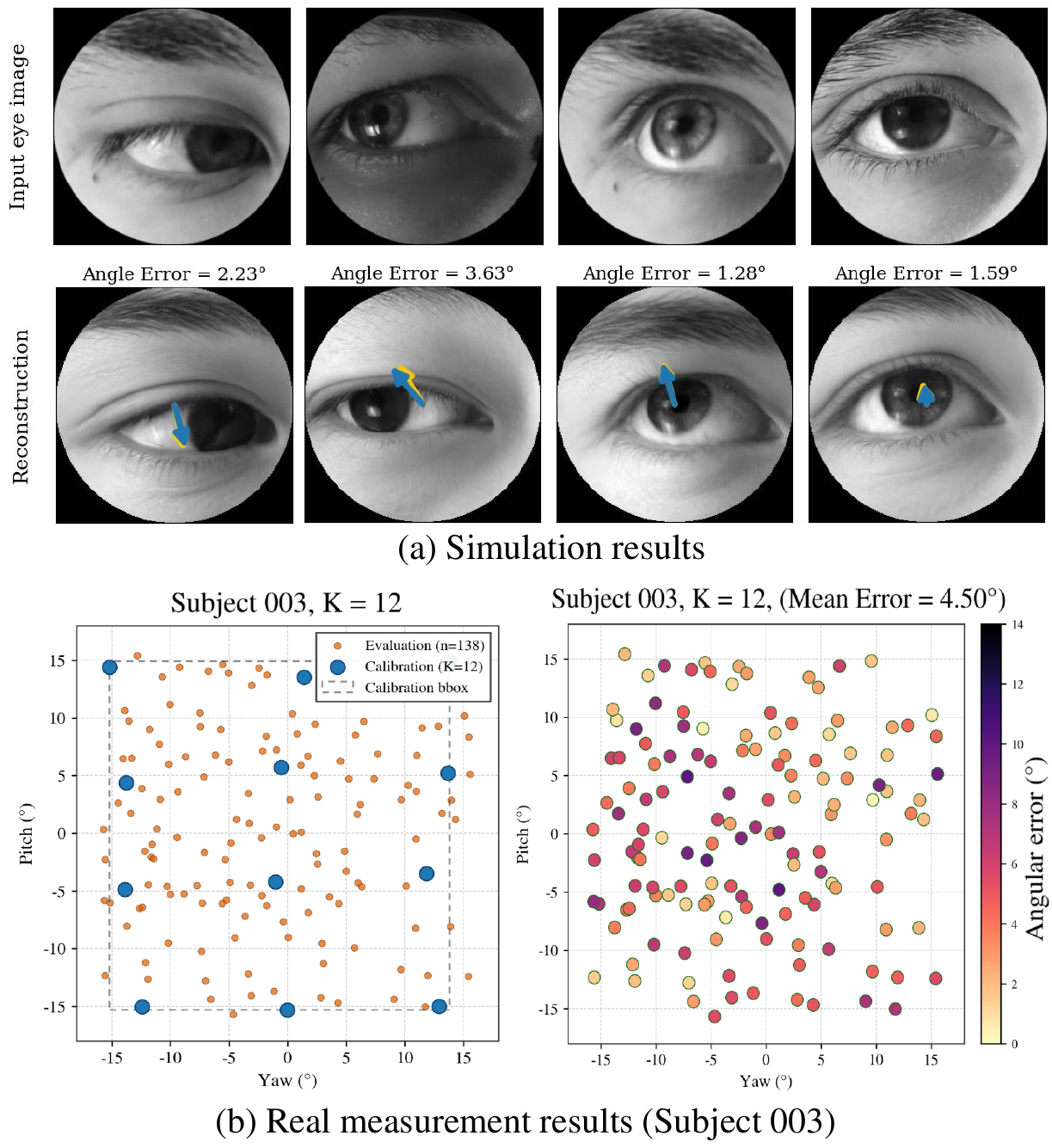}
    \else
        \includegraphics[width=1\linewidth]{figures/result.pdf}
    \fi
   \caption{\textbf{Evaluation results in simulation and real-world hardware.} (a) Representative gaze estimation results from the cross-subject simulation. The top row shows the input eye patches, while the bottom row displays the corresponding reconstructions overlaid with ground-truth (yellow) and predicted (blue) gaze vectors. (b) Real hardware measurement results for Subject 003. The left plot illustrates the spatial distribution of $n=138$ evaluation points relative to $K=12$ calibration points. The right heatmap visualizes the per-point angular error across the field of view, spanning approximately $\pm15^{\circ}$ in pitch and yaw, achieving a mean error of $4.50^{\circ}$ after subject-specific fine-tuning.}
  \Description[Simulation gaze vectors and real-world angular error distribution]{Top is Representative gaze estimation results from the cross-subject simulation. Bottom is Real hardware measurement results for Subject 003}
   \label{fig:result}
\end{figure}

\begin{figure*}[t]
  \centering
   \includegraphics[width=0.82\linewidth]{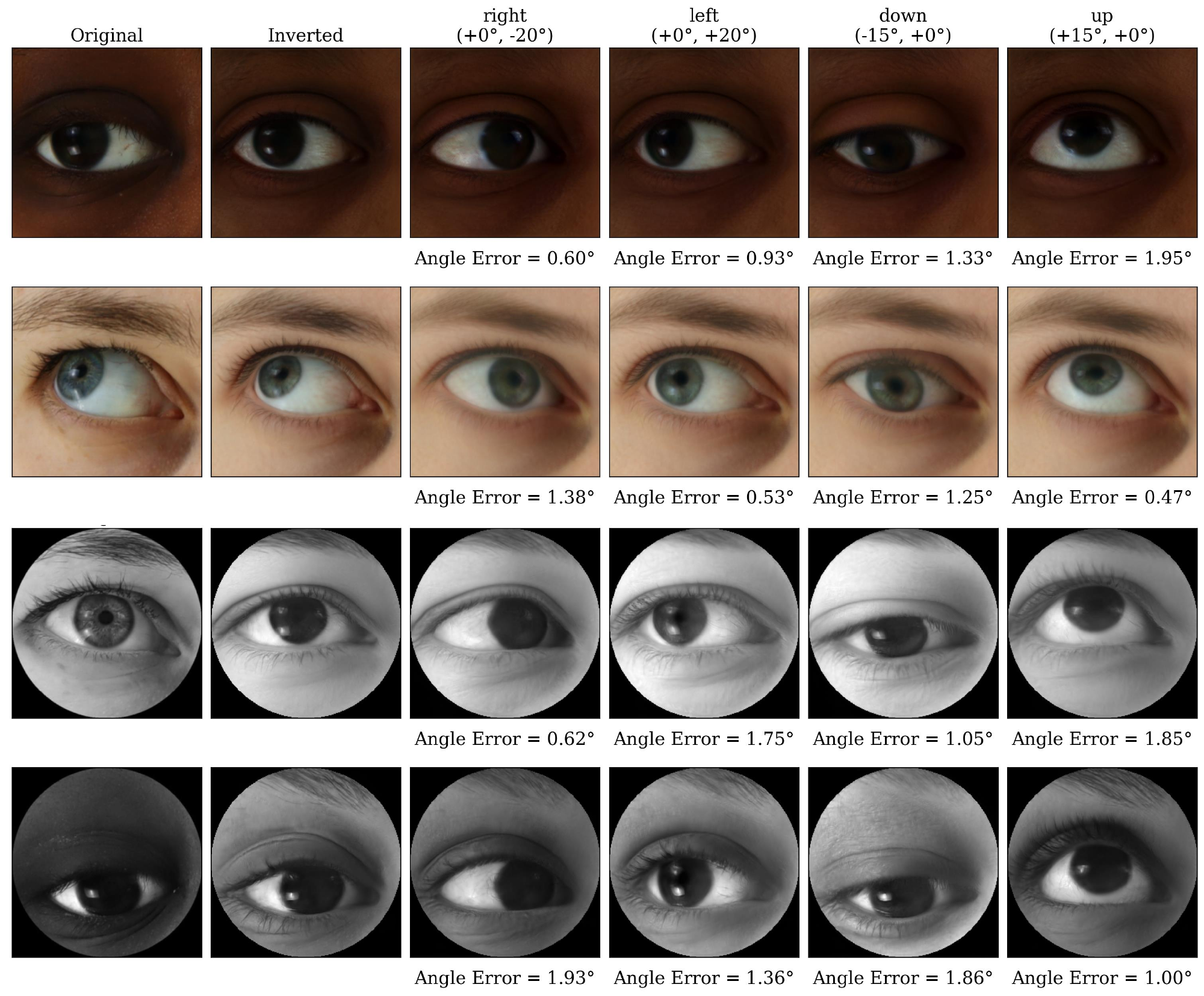}
\caption{\textbf{Gaze steering example.} Original / Inverted shows the source eye and its latent inversion from the simulated 16-dimensional sensor measurements. We steer gaze to right $(0^\circ,-20^\circ)$, left $(0^\circ,+20^\circ)$, down $(-15^\circ,0^\circ)$, and up $(+15^\circ,0^\circ)$. Grayscale results show that the compact latent space preserves steerable gaze information while varying the appearance, thus maintaining privacy, whereas color results use full image-space inversion for higher fidelity in the identity of the subject.}  \Description {Image of gaze steering.}
   \label{fig:steer}
\end{figure*}

To thoroughly evaluate the performance of our all-optical latent feature acquisition system, we design a two-stage assessment pipeline followed by a dedicated analysis of system latency. The first stage consists of a fully simulation-based evaluation, where the optical encoder $\encodersoft$ and its associated binary masks are optimized and trained alongside the decoder $\decoder$ using our curated dataset of eye images, as discussed in Section~\ref{sec:model}. The second stage transitions to real-world deployment through human-subject fine-tuning. We fabricated the optimized masks from the first stage and assemble them with MLA and phototransistors. The system processes direct optical input from human eyes; for each participant, we apply a rapid subject-specific calibration to fine-tune the weights of $\decoder$, yielding the final inference $\decodersub$. This two-step approach allows us to bridge the gap between theoretical optical design and practical low-latency gaze tracking.

\subsection{Fully Simulated Evaluation}
\label{sec:simulated_evaluation}

We evaluate whether gaze can be estimated from a compact set of learned optical measurements by comparing our method with image-based baselines under controlled sensing and computation budgets. 
The goal is to test whether task-optimized optical sensing can preserve gaze-relevant information while reducing both the number of acquired measurements and the complexity of the downstream predictor.

All models are trained and evaluated on monocular ETH-XGaze eye patches~\cite{Zhang2020ETHXGaze}, using the subject split described in Section~\ref{sec:dataset}, 
unless otherwise specified, we train using the original training set together with the generative augmentation strategy described in Section~\ref{sec:aug_gaze}. 
For the full-resolution ResNet18 baseline, this augmentation improves the angular error from $3.33^\circ$ to $3.08^\circ$, suggesting that diversifying periocular appearance while preserving gaze labels improves cross-subject generalization.
We compare against representative convolutional backbones used in modern gaze-estimation pipelines, including ResNet-family models~\cite{park2020towards,cheng2022puregaze,athavale2022one}, as well as MobileNetV3~\cite{howard2017mobilenets} and EfficientNet-B0~\cite{tan2019efficientnet}. 
These architectures are relevant baselines because they are commonly used as compact encoders or bottlenecks for edge-oriented vision systems. 
Image-based baselines operate either on $16{\times}16$ or $256{\times}256$ eye patches. 
In contrast, our method does not acquire a dense image; it estimates gaze from 16 learned optical measurements captured by a $4{\times}4$ phototransistor array and decoded by a lightweight MLP trained with latent-space distillation, as described in Section~\ref{sec:optical_latent_encoder}.

Fig~\ref{fig:result}(a) qualitatively illustrates this sensing regime. 
The learned measurements preserve enough gaze information to recover a representative eye sample with a similar gaze direction, but the reconstruction is not intended to preserve the subject's identity or detailed appearance. 
Instead, it corresponds to a plausible eye within the generative model distribution that matches the inferred gaze. 
This is desirable for privacy-preserving gaze sensing, since eye-tracking data can reveal sensitive personal information~\cite{kroger2020gaze,liebling2014privacy}, whereas our sensing pipeline avoids capturing or reconstructing a faithful biometric eye image (Figure~\ref{fig:steer}).

Table~\ref{tab:compactness_accuracy} summarizes the compactness--accuracy tradeoff. 
Our method achieves $6.47^\circ$ angular error using only 16 measurements, 0.244M parameters, and 0.49 MFLOPs. 
Compared with the $16{\times}16$ ResNet18 baseline, it reduces sensing dimensionality by $16\times$, parameters by approximately $48\times$, and predictor computation by approximately $573\times$, with a moderate error increase from $5.54^\circ$ to $6.47^\circ$. 
Compared with the full-resolution $256{\times}256$ ResNet18 baseline, it reduces sensing dimensionality by $4096\times$ and predictor computation by approximately $9300\times$.

The compact-backbone comparisons show that the proposed approach is not simply a smaller neural network. 
At $16{\times}16$, MobileNetV3 obtains $6.87^\circ$ error with 1.503M parameters and 4.27M FLOPs, while our method achieves lower error with approximately $6.2\times$ fewer parameters and $8.7\times$ fewer FLOPs. 
EfficientNet-B0 improves the error to $5.87^\circ$, but requires approximately $21\times$ more parameters and $40\times$ more FLOPs. 
Thus, even aggressively downsampled image-based baselines still require forming and processing an eye image, whereas our system directly acquires a small set of task-optimized optical measurements.

We further examined whether accuracy depends on the gaze range by
evaluating the same model on test subsets restricted to increasingly
wide ranges. The error varies by less than $0.3^\circ$ across these
subsets, from $6.17^\circ$ within $\pm15^\circ$---the range covered by
the hardware evaluation in Sec.~\ref{sec:human_eval}---to $6.47^\circ$
over the full test span, indicating that accuracy is limited by the
sensing budget rather than by how far the eye is rotated.

Overall, these results show that the proposed system targets an extreme low-measurement regime rather than the unconstrained accuracy regime of dense image-based gaze estimators. 
By combining learned optical sensing with latent-space supervision, the system provides a compact, identity-obfuscating pipeline for low-power and low-latency gaze estimation.
\begin{table}[t]
\centering
\caption{
Compactness--accuracy comparison on ETH-XGaze eye patches. 
Lower angular error is better. 
MobileNetV3 and EfficientNet-B0 are included as representative edge-oriented image backbones. 
Params and FLOPs refer to the digital predictor after sensing.
}
\label{tab:compactness_accuracy}
\ifarxiv{
\begin{tabular}{lcrrr}
\toprule
Method & Output & Params (M) & FLOPs (M) & 
\begin{tabular}{c}Angular\\error ($^\circ$) $\downarrow$\end{tabular} \\
\midrule
ResNet18 & $16{\times}16$ img. & 11.697 & 280.10 & 5.54 \\
ResNet18 & $256{\times}256$ img. & 11.697 & 4546 & 3.08 \\
\midrule
MobileNetV3 & $16{\times}16$ img. & 1.503 & 4.27 & 6.87 \\
MobileNetV3 & $256{\times}256$ img. & 1.503 & 143.04 & 3.11 \\
\midrule
EfficientNet-B0 & $16{\times}16$ img. & 5.124 & 19.51 & 5.87 \\
EfficientNet-B0 & $256{\times}256$ img. & 5.124 & 1029.00 &  \textbf{2.96} \\
\midrule
Ours & 16 meas. & \textbf{0.244} & \textbf{0.49} & 6.47 \\
\bottomrule
\end{tabular}
}
\else{
\scriptsize
\setlength{\tabcolsep}{2.5pt}
\resizebox{\columnwidth}{!}{
\begin{tabular}{lcrrr}
\toprule
Method & Output & Params (M) & FLOPs (M) & 
\begin{tabular}{c}Angular\\error ($^\circ$) $\downarrow$\end{tabular} \\
\midrule
ResNet18 & $16{\times}16$ img. & 11.697 & 280.10 & 5.54 \\
ResNet18 & $256{\times}256$ img. & 11.697 & 4546 & 3.08 \\
\midrule
MobileNetV3 & $16{\times}16$ img. & 1.503 & 4.27 & 6.87 \\
MobileNetV3 & $256{\times}256$ img. & 1.503 & 143.04 & 3.11 \\
\midrule
EfficientNet-B0 & $16{\times}16$ img. & 5.124 & 19.51 & 5.87 \\
EfficientNet-B0 & $256{\times}256$ img. & 5.124 & 1029.00 &  \textbf{2.96} \\
\midrule
Ours & 16 meas. & \textbf{0.244} & \textbf{0.49} & 6.47 \\
\bottomrule
\end{tabular}}
}
\fi
\end{table}

\subsection{Gaze Steering Evaluation}
To assess identity and appearance preservation in the gaze-steering operator, we report a cyclic round-trip evaluation. Starting from the source reconstruction, we first steer the gaze to a target direction and then steer it back to the original direction, comparing the cycled image with the source reconstruction. This provides a meaningful reference for image-quality metrics such as PSNR, SSIM, and LPIPS, since the forward edit alone has no ground-truth target image. The forward and backward angular errors, $1.18^\circ$ and $1.23^\circ$, confirm accurate steering in both directions, while the photometric scores, PSNR $29.02$ dB, SSIM $0.879$, and LPIPS $0.125$, indicate that appearance is well preserved through the full round trip. Qualitative examples are shown in Fig.~\ref{fig:steer}.

We use gaze steering as a lightweight latent-space editing module rather than a dedicated steering network. Given the source and target gaze directions, a small MLP predicts a gated offset in $\fullwplusedit$, enabling controlled gaze edits with minimal architectural complexity. Prior work used gaze steering to synthesize adaptation data and reported small improvement in gaze-estimation accuracy \cite{yu2019improving}; however, we did not observe a comparable adaptation gain in our setting. Instead, our results highlight gaze steering as a practical editing tool for graphics applications, requiring only a compact MLP rather than a specialized steering architecture.

\subsection{Evaluation with Human Subjects}
\label{sec:human_eval}
To evaluate the real-world performance of our system and bridge the simulation-to-reality gap, we conducted a series of experiments with human subjects using our hardware prototype as shown on the right of Fig.~\ref{fig:teaser}.

During acquisition, subjects stabilized their heads and observed a monitor through the beam splitter. Calibration targets were randomly presented across the screen to approximately uniformly cover the gaze range, and the known screen geometry was used to compute the ground-truth gaze direction. No eye images were captured during evaluation. Illumination was provided by standard broadband room lighting. For each target, the FPGA collected continuous raw intensity signals for less than one second at $900~\mathrm{Hz}$. The phototransistor outputs were normalized using white and black reference measurements, filtered for electronic noise and  to form the final 16-dimensional latent measurements.

\begin{table}[t]
\centering
\caption{Real-measurement calibration results across 12 subjects. $K$ denotes the number of subject-specific calibration points used to fine-tune the decoder; MAE is the mean angular error.}
\label{tab:real_calibration}
\small
\begin{tabular}{@{}lccc@{}}
\toprule
$K$ & 9 & 12 & 15 \\
\midrule
MAE ($^\circ$) $\downarrow$ & 6.55 & 5.96 & 5.92 \\
\bottomrule
\end{tabular}
\end{table}

\begin{figure*}[ht]
  \centering
   \includegraphics[width=0.8\linewidth]{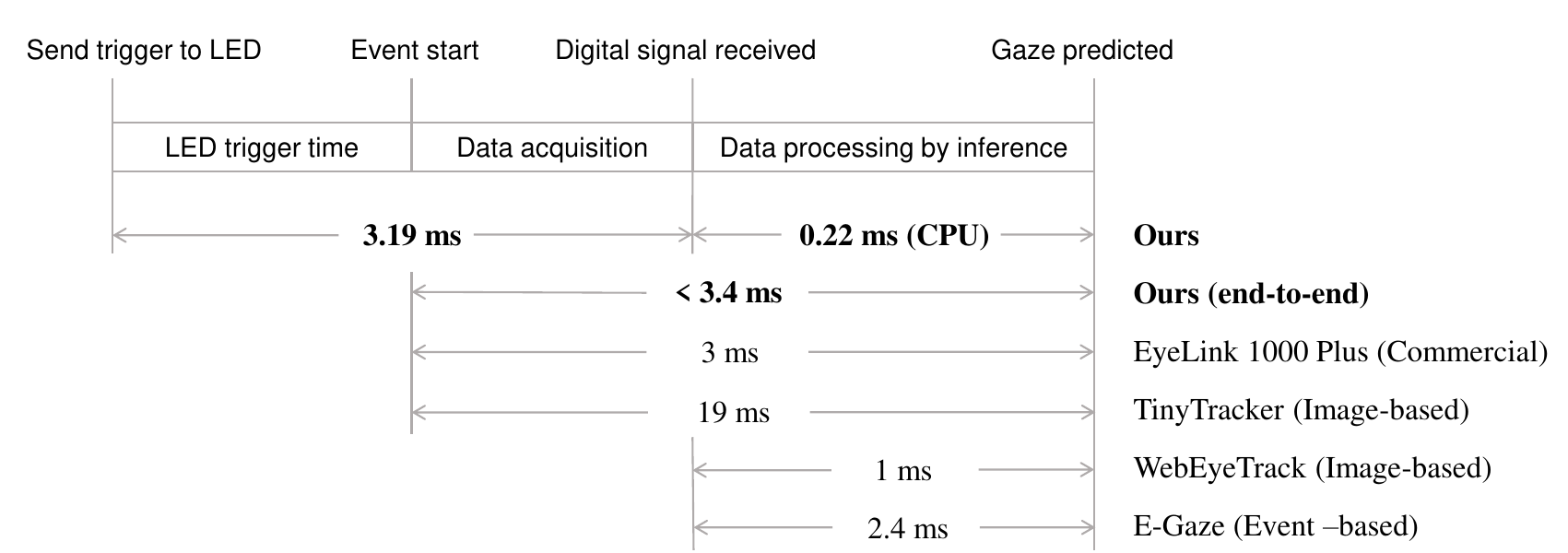}
    \caption{\textbf{Timing diagram and latency comparison.} Our end-to-end pipeline is measured from the initial LED trigger through physical acquisition (3.19 ms) to final CPU inference (0.22 ms), totaling less than 3.4 ms. This performance is compared against state-of-the-art commercial~\cite{eyelink1000plus}, image-based~\cite{10325167,davalos2025webeyetrack}, and event-based systems~\cite{li2024egaze}, demonstrating that our all-optical latent sensing approach matches specialized high-speed trackers while significantly outperforming conventional mobile edge vision systems.}
    \Description{Timing diagram and comparative bar chart of system latency}
   \label{fig:latency}
\end{figure*}

We conducted experiments with 12 human subjects using the hardware prototype. Data were collected under ordinary indoor lighting using subjects of different genders and ethnicities. The evaluation naturally included variations in eye color, contact lenses, eye makeup, and small involuntary head movements. Subjects self-initiated each measurement once they believed they had stably fixated on the displayed target; trials affected by obvious blinks or large unintended movements were simply repeated. For each subject, we captured 150 high-quality data points, with a small number of remaining outliers automatically rejected based on deviations from adjacent samples. We selected $K$ data points from this set for rapid subject-specific calibration of $\mathcal{P}_{sub}$. These points were distributed across the entire gaze area to ensure comprehensive coverage, reporting results for $K=9, 12, \text{ and } 15$ in Table~\ref{tab:real_calibration}. Notably, with $K=12$ calibration points, the system already achieves a mean angular error of $5.96^{\circ}$, with some subjects reaching a minimum error of $4.5^{\circ}$. Fig.~\ref{fig:result}(b) illustrates this process for a representative participant (Subject 003), showing the spatial distribution of the $K=12$ calibration points alongside the evaluation points (left), as well as the resulting gaze estimation errors across the field of view (right). 

Two geometric factors are compensated by the per-subject calibration.
Without a corneal glint as reference, residual sensor--eye
misalignment is geometrically ambiguous from eye
rotation \cite{niehorster2020impact}, which glint-based methods
mitigates by construction \cite{Guestrin2006}; and the optical axis
observed by appearance-based sensing differs from the visual axis by
the kappa angle, ${\sim}5^\circ$, individual-specific, and learnable
only on average across subjects. Both are constant within a session,
so twelve calibration points recover most of the error, with little
gain beyond. The $\pm15^\circ$ range is determined by the geometry of our proof-of-concept
setup, not the encoding principle: the same encoder covers the full
test span with nearly uniform error
(Sec.~\ref{sec:simulated_evaluation}), making the calibrated
$5.96^\circ$ comparable to the matched-range simulation error of
$6.17^\circ$. Since foveal vision spans only
${\sim}2^\circ$~\cite{patney2016towards}, this accuracy supports
region-level gaze applications---attention, dwell, and coarse
selection---rather than the sub-degree regime of clinical work or
precise pointing; at $6.3^\circ$, LiGaze distinguishes 4, 9, and 16
screen regions with $100\%$, $99.5\%$, and $91.7\%$
accuracy~\cite{10.1145/3131672.3131682}.

\subsection{System Speed and Latency}
We evaluate the end-to-end latency of the proposed system, including optical sensing, signal acquisition, and computational inference. All latency is measured over 2000 runs after 50 warm-up iterations.

\paragraph{Measurement Setup.}
To measure the physical sensing latency, we use an LED as a controlled optical trigger. A digital signal is sent to turn on the LED, and the resulting change in measured intensity is recorded by the phototransistor array and read by the PC. The time difference between the trigger signal and the recorded intensity change reflects the sensing and acquisition latency.

\paragraph{Optical Sensing Latency.}
Across repeated measurements, the average latency from LED trigger to PC readout is $3.19~\mathrm{ms}$. This includes phototransistor response, analog-to-digital conversion, FPGA readout, and data transfer. The sensing pipeline operates at $900~\mathrm{Hz}$, corresponding to a frame interval of approximately $1.11~\mathrm{ms}$.

\paragraph{Inference Latency.}
Inference is evaluated on a workstation equipped with an AMD EPYC 7763 CPU and an NVIDIA RTX A4500 GPU. On a CPU, the forward pass takes approximately $0.22~\mathrm{ms}$. On a GPU, the network inference itself requires $0.17~\mathrm{ms}$. When accounting for data transfer between CPU and GPU, an additional overhead is observed. The total GPU-based inference latency, including data transfer, is approximately $0.68~\mathrm{ms}$.

\paragraph{End-to-End Latency.}
Combining sensing and inference, the total system latency is approximately $3.4~\mathrm{ms}$ on CPU and $3.87~\mathrm{ms}$ on GPU. This demonstrates that the proposed system achieves sub-$5~\mathrm{ms}$ latency, significantly outperforming conventional camera-based gaze tracking systems that typically operate at tens of milliseconds due to image acquisition and processing overhead. A comparison with representative systems is provided in Figure~\ref{fig:latency}. The temporal breakdown of our trigger-to-output pipeline and its comparison with state-of-the-art image-based and event-based trackers are further illustrated in Fig.~\ref{fig:latency}.

It is important to note that latency definitions vary across prior work. For example, in recent event-based gaze tracking systems such as~\cite{li2024egaze}, latency is defined as \emph{“the time required to estimate gaze from an event set”}, which corresponds to processing time after events have been accumulated. As a result, this definition does not explicitly account for the event acquisition process itself. In practice, event generation depends on brightness changes and can become significantly slower under low-light or low-contrast conditions, introducing additional sensing delay that is not reflected in the reported latency.

While high-end commercial systems such as the EyeLink 1000 Plus~\cite{eyelink1000plus} report an end-to-end latency of approximately $3~\mathrm{ms}$, our measured sensing latency of $3.19~\mathrm{ms}$ includes the inherent activation delay of the LED trigger used in our setup. Since this trigger introduces additional delay in the measurement loop, the reported latency should be considered an upper bound. Accounting for this measurement overhead, the actual latency of the optical sensing-to-inference pipeline is expected to be lower, suggesting that the proposed system can match or potentially exceed the temporal performance of specialized hardware while maintaining a fully passive and compact architecture.

\section{Discussion and Outlook}\label{sec:conclusion}

In this work, we have demonstrated an all-optical latent feature acquisition system that achieves low-latency gaze tracking by shifting the computational burden of feature extraction into the optical domain. By utilizing a task-driven optical encoding, we bypass the conventional image-formation bottleneck, directly capturing a compact 16-dimensional representation of the eye.

The primary advantage of the proposed pipeline is its extreme efficiency on both the sensing and computational sides. As shown in our latency analysis, the physical acquisition takes only 3.19 ms, while the lightweight digital predictor requires a mere 0.22 ms on a CPU. This combined performance results in a total system latency of less than 3.4 ms, successfully meeting the critical 5 ms threshold required for immersive XR applications. Unlike image-based systems that are limited by CMOS frame rates and massive data readout, our sensing-as-inference paradigm enables a high sampling rate of 900~Hz with minimal data volume.

Our evaluation results highlight the effectiveness of the two-stage training strategy. In a fully simulated cross-subject setting, the system achieves a mean angular error of 6.47$^\circ$ without individual fine-tuning. Our real-world experiments demonstrate that rapid subject-specific fine-tuning successfully bridges the simulation-to-reality gap. With only 12 to 15 calibration points, the per-subject mean angular error is reduced to below 6$^\circ$. This confirms that the learned latent features are not only robust to physical fabrication tolerances but are also expressive enough to capture gaze-relevant information across different human subjects.

While our current hardware serves as a proof-of-concept prototype, it
establishes a blueprint for more efficient commercial
implementations. The massive reduction in measurement
dimensionality---from a $256\times256$ image to just 16
scalars---represents a $4096\times$ decrease in sensing data. This
reduction directly translates to lower power consumption in data
transmission and processing, which is vital for battery-operated
mobile XR glasses. State-of-the-art camera-based gaze trackers
achieve substantially higher accuracy, but rely on richer sensing:
high-resolution eye images, active infrared illumination, larger
input regions (both eyes or the full face), and more extensive
calibration, together with considerably more computational
processing~\cite{Zhang2020ETHXGaze,kellnhofer2019gaze360,cheng2022puregaze}. Our work instead explores a different point on the accuracy-latency-compute trade-off. Consequently, direct comparison of angular accuracy alone does not fully reflect the different design objectives of these systems.
Future implementations could further improve accuracy and integration
through better optical fabrication, alignment, and miniaturization
while preserving the core sensing paradigm. Within the regime
identified above, such systems are particularly attractive for
latency- and power-constrained applications---smart glasses, near-eye
displays, attention monitoring, coarse gaze interaction, and foveated
rendering with conservatively padded foveal regions---where extremely
low end-to-end latency and compact passive hardware matter more than
the highest attainable gaze accuracy.

\bibliographystyle{unsrtnat}
\bibliography{bibliography}

\appendix
\newpage
\begin{center}
    {\LARGE \textbf{Supplementary Material}}
\end{center}

\section{Latent Space Gaze Steering}

\begin{figure*}
  \centering
   \includegraphics[width=0.9\linewidth]{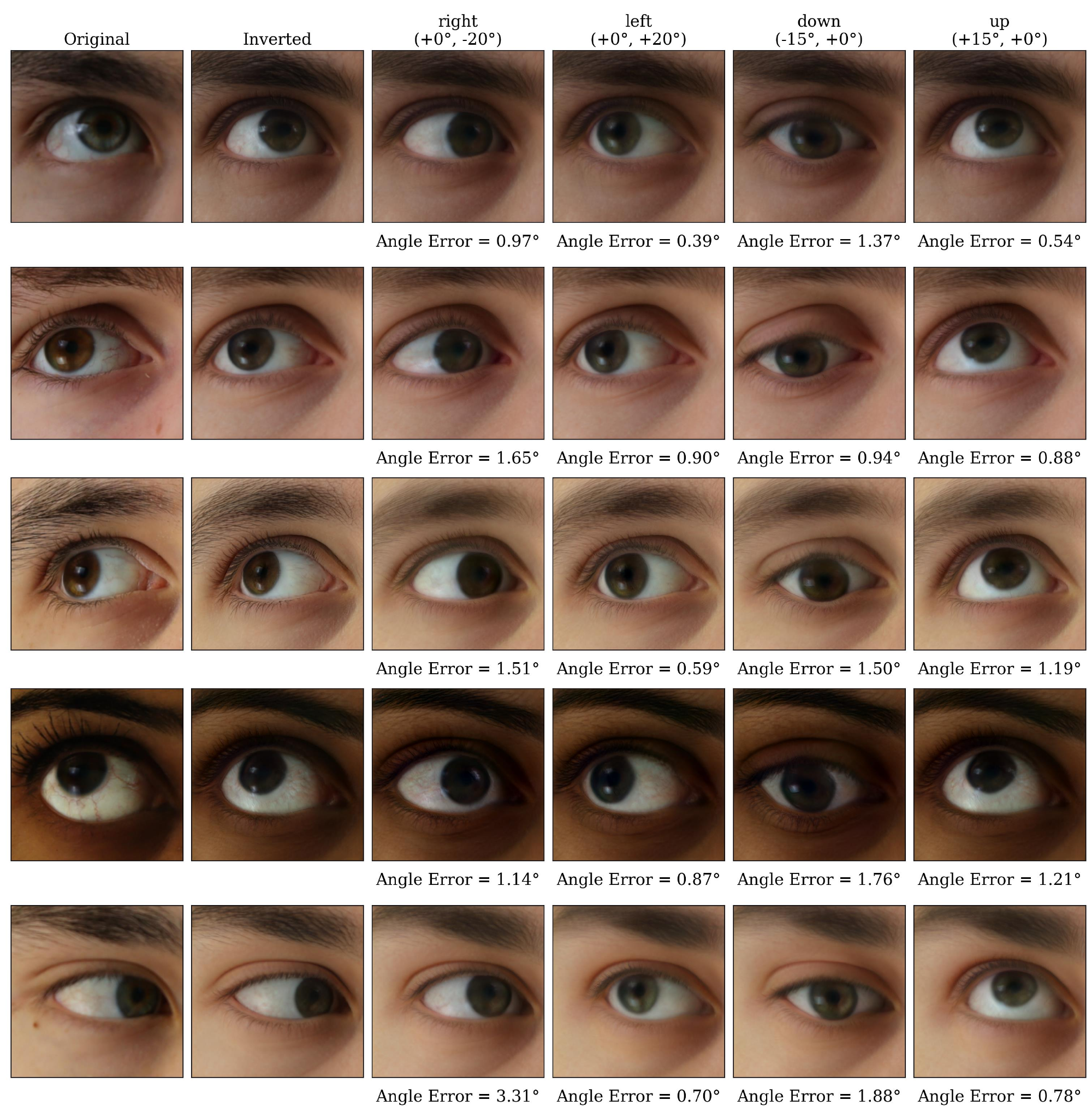}
\caption{Results for gaze steering in latent space.}
  \Description
  {Results for gaze steering in latent space.}
   \label{fig:gaze_steer}
\end{figure*}

We provide additional qualitative results for gaze steering to demonstrate the semantic richness of the learned latent representations. By applying a control network $C_{\theta}$ to predict a residual displacement in the $\mathcal{W}^{+}$ space, we can manipulate the gaze direction of a captured eye while preserving the subject's unique periocular identity and appearance.  As illustrated in Fig.~\ref{fig:gaze_steer}, our system successfully performs large-angle redirections (up to $\pm 20^{\circ}$) across diverse subjects. The steered images maintain high anatomical plausibility, with the control network effectively re-orienting the pupil and iris while adjusting the surrounding eyelid geometry to match the new gaze target. The low angular errors reported between the target gaze and the re-estimated gaze from the steered images further confirm that the latent updates are geometrically consistent.

\section{Dataset Details}
\label{sec:dataset_details}
Our dataset is derived from the ETH-XGaze dataset
\citesupp{Zhang2020ETHXGaze} that contains high-resolution RGB images
under regular room illumination. Since our system is based on a sensor
zoomed in on an individual eye (similar to what may be found in a
near-eye display), the dataset needs to be adjusted for this setting.
We specifically select images from
cameras positioned frontally relative to the subjects. To ensure
consistency across binocular data, we normalize both the left and
right eye images \citesupp{Gaze_normal, Zhang2020ETHXGaze}. Left eye
images are horizontally flipped, and the $x$-component of their
corresponding gaze vectors is negated to maintain a unified coordinate
system. 

The data processing pipeline follows a multi-stage cropping and augmentation strategy:
\begin{itemize}
    \item \textbf{Initial Normalization:} Images are first cropped and normalized to a resolution of $288 \times 288$ pixels, representing a physical area of $45 \times 45~\mathrm{mm}$.
    \item \textbf{Augmentation and Robustness:} During training, we apply random translations and scaling to these patches to extract the final input eye patches $I \in \mathbb{R}^{256 \times 256}$, corresponding to a $40 \times 40~\mathrm{mm}$ region. This augmentation process is specifically designed to mimic potential optical alignment errors and mechanical tolerances in the physical prototype, ensuring the network remains robust to slight shifts in the microlens array relative to the eye.
    \item \textbf{Artifact Mitigation:} To eliminate corner artifacts inherent to microlens array imaging, a circular mask is applied to the final eye patches.
\end{itemize}

Each eye patch is annotated with a ground-truth 3D gaze vector $\mathbf{g} \in \mathbb{R}^3$ and a validity label $v \in \{0, 1\}$, where $v=0$ indicates invalid inputs such as closed eyes, offset irises, or non-eye patches. These labels were initially manually annotated on a subset to train a classifier, which then automatically labeled the remaining corpus.

\begin{figure*}
  \centering
   \includegraphics[width=1\linewidth]{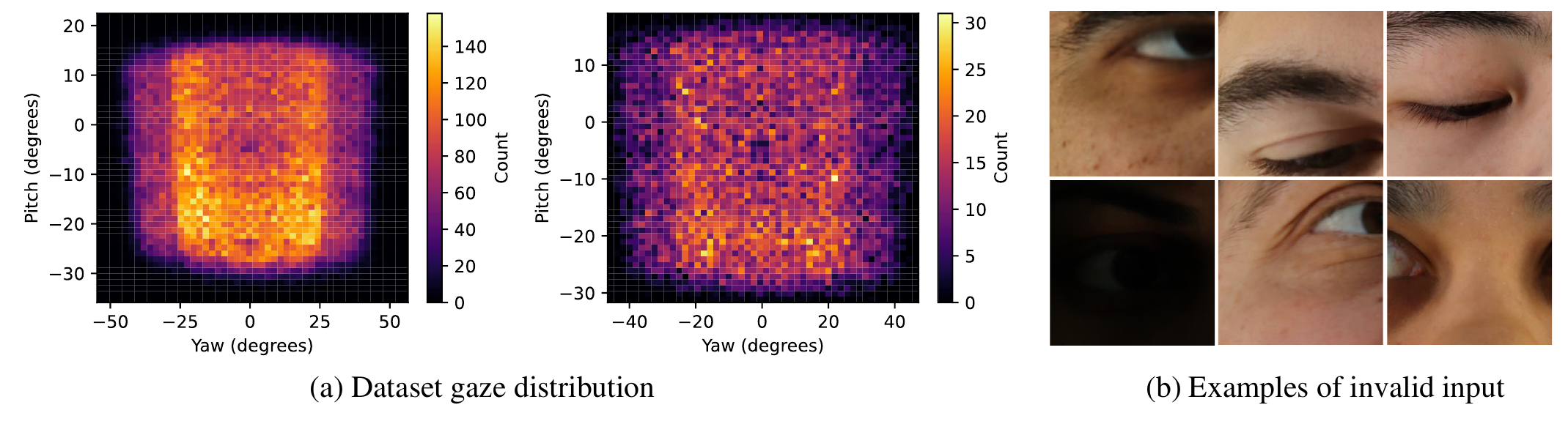}
\caption{\textbf{Gaze Distribution and Validity Classification.} (a) Heatmaps showing the distribution of 3D gaze vectors for the training set (68 subjects, left) and the test set (12 subjects, right). The dataset covers a wide field-of-view with pitch ranging from $-35.87^\circ$ to $22.45^\circ$ and yaw from $-54.73^\circ$ to $56.5^\circ$. (b) Representative examples of invalid eye patches ($v=0$) identified by our automatic classifier, including closed eyes, extreme iris offsets, and non-eye regions, which are used to improve system robustness during real-world operation.}
  \Description[Gaze distribution heatmaps and examples of invalid eye images]
  {A figure consisting of three parts. Part (a) shows two heatmaps representing the density of gaze directions in degrees for the training and test sets; the patterns show a broad, consistent coverage across a wide horizontal and vertical range. Part (b) displays a grid of circular eye patches that are blurry, occluded, or do not contain a centered eye, illustrating the types of data points the system must recognize as invalid.}
   \label{fig:dataset}
\end{figure*}

Our processed dataset consists of 68 subjects for training and 12 subjects for testing. The training set contains 142,616 images (128,379 valid), while the test set contains 25,504 images (23,828 valid). The gaze coverage spans a pitch range of $-35.87^\circ$ to $22.45^\circ$ and a yaw range of $-54.73^\circ$ to $56.5^\circ$. Figure~\ref{fig:dataset} illustrates the distribution of gaze directions and examples of the automatic validity classification.

\section{Details of Identity-Varying Augmentation via Generative Inpainting}
\label{sec:details_aug}
\begin{figure*}
  \centering
   \includegraphics[width=0.9\linewidth]{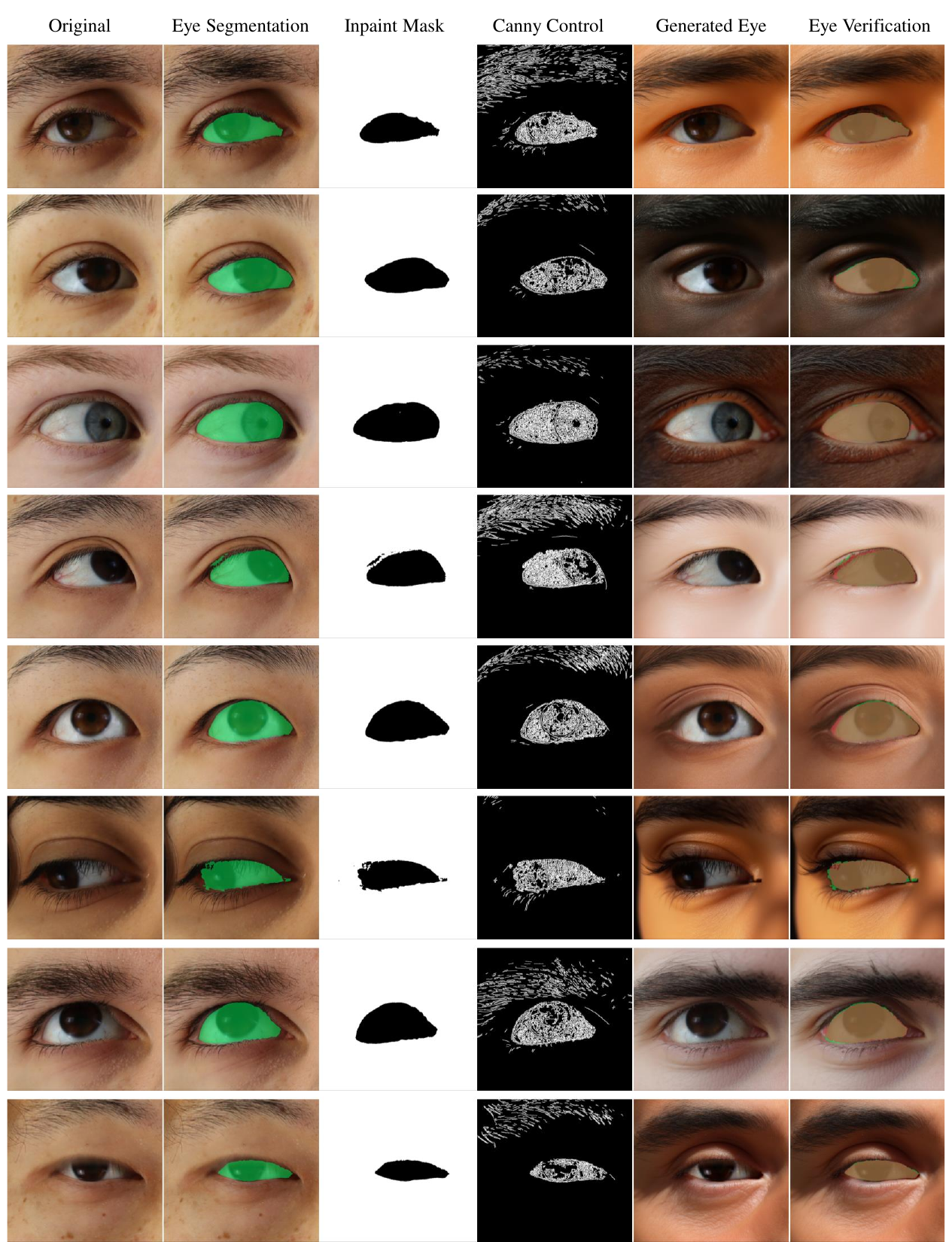}
\caption{\textbf{Identity-varying augmentation via generative
inpainting.} From left to right: original eye crop, SAM3 eye segmentation, inpainting mask, Canny-edge ControlNet input, Flux-generated eye conditioned on the mask, Canny edges, and a diversity prompt, and SAM3 re-segmentation of the generated eye overlaid on the original mask for consistency checking. Samples failing the area/IoU check are replaced by the original crop, so the gaze label always remains valid.}
  \Description[Identity-varying augmentation via generative
inpainting]
  {From left to right: original eye crop, SAM3 eye segmentation, inpainting mask, Canny-edge ControlNet input, Flux-generated eye conditioned on the mask, Canny edges, and a diversity prompt, and SAM3 re-segmentation of the generated eye overlaid on the original mask for consistency checking. \red{Samples failing the area/IoU check are replaced by the original crop, so the gaze label always remains valid.}}
   \label{fig:augmentation_sup}
\end{figure*}

To expand our dataset we perfomed a generative based augmentation, we know that appearance-based gaze estimation models trained on datasets such as ETH-XGaze \citesupp{Zhang2020ETHXGaze} are prone to identity bias: each
subject contributes thousands of images across varying gaze directions, encouraging the model to learn identity-specific shortcuts —
skin tone, eye shape, eyebrow structure — rather than gaze-relevant features. This is compounded by the limited demographic diversity
typical of laboratory-collected datasets, where certain ethnicities, age groups, and facial characteristics are underrepresented \citesupp{akgul2026gazeFairness}.

To address this, we introduce an augmentation scheme that leverages the Flux text-to-image diffusion model \citesupp{flux2024} to expand the range of subject appearances associated with each gaze label. The motivation is that appearance-based gaze estimators may rely not only on the iris, pupil, and sclera, but also on surrounding periocular cues such as skin texture, eyebrow shape, facial structure, and illumination. By varying these appearance-related factors while encouraging the synthesized images to preserve the original eye configuration, the proposed augmentation exposes the model to broader identity and contextual variation without deliberately changing the annotated gaze direction.

Our pipeline operates as follows. Given an eye crop with its associated gaze vector, we first obtain the eye region using SAM3
\citesupp{carion2025sam3segmentconcepts}. We then construct an inverted inpainting mask that preserves the eye pixels and marks the surrounding
context for regeneration. Canny edges \citesupp{4767851} extracted from the original image serve as a structural control signal via
ControlNet \citesupp{zhang2023adding}, maintaining anatomical plausibility of the generated context. Flux then inpaints the masked region
conditioned on a text prompt sampled from a curated pool of descriptions spanning diverse skin tones, age groups, makeup variations, and lighting conditions. Each generated sample undergoes two quality checks: a CLIP-based \citesupp{radford2021learning}
prompt alignment score to ensure semantic fidelity, and an eye consistency verification that re-segments the eye in the output and
requires both area preservation (within 10\%) and spatial overlap (IoU $\geq$ 0.9) with the original. Samples failing either check are regenerated with a different seed, up to three attempts.

This approach produces augmented images that are geometrically faithful, the eye region and its gaze label are untouched, while
exhibiting substantial variation in the surrounding appearance. By expanding each training sample into multiple identity-diverse variants, we reduce the model's reliance on identity-correlated features and improve generalization. Examples of the augmentation process are shown in Figure~\ref{fig:augmentation_sup}.

\paragraph{Implementation details.}
Before edge extraction the crop is locally contrast-equalized (CLAHE, clip limit
$3$, $8\times8$ tiles), and the Canny thresholds are lowered inside a dilated
neighborhood of the eye so that lid and iris contours dominate the structural
signal; edge components smaller than $30$ pixels are discarded. Seeds are derived
deterministically from the sample index, so the corpus can be regenerated
exactly. Table~\ref{tab:aug_config} lists the remaining settings.
  
\begin{table}[h]
\centering
\small
\caption{Augmentation configuration.}
\label{tab:aug_config} 
\begin{tabular}{ll}
\toprule
Segmentation           & SAM3, text prompt \emph{``eye''} \\
Inpainting             & FLUX.1-dev + Canny ControlNet, \texttt{bf16} \\
Semantic check         & CLIP ViT-L/14, score $\geq 23$ \\
Generation resolution  & $768\times768$, stored at crop resolution \\
Steps / guidance       & $40$ / $4.5$ \\
Strength / ControlNet scale & $0.97$ / $0.7$ \\
Attempts per image     & up to $3$, distinct seeds \\
Variants per source image & $1$ \\
\bottomrule
\end{tabular}
\end{table}

\paragraph{Prompt pool.}
All prompts share the template \emph{``raw photo, close-up of a person's face
around the eye, \{variation\}, shot on DSLR''}. Because every crop is right-eye
oriented, each one carries the suffix \emph{``inner eye corner and nose bridge on
the right side, eyebrow sweeping from right to left''}, which keeps the 
synthesized periocular anatomy consistent with the crop geometry.
Table~\ref{tab:aug_prompts} summarizes the pool; representative variations are:
\begin{itemize}\itemsep1pt \parskip0pt 
\item \emph{``thin metal rectangular glasses frame over the brow, fair skin on cheek, indoor lighting''}
\item \emph{``natural no-makeup look on forehead and cheek, visible pores, soft natural lighting''}
\item \emph{``crow's feet wrinkles on temple skin, pale cheek, natural lighting''}
\item \emph{``pale skin, cool blue-tinted indoor fluorescent light''}
\item \emph{``slight sweat on brow, flushed cheeks, outdoor exertion''}
\end{itemize}

\begin{table}[h]
\centering
\small
\caption{Composition of the prompt pool.}
\label{tab:aug_prompts} 
\begin{tabular}{lr}
\toprule
Skin tone and ethnicity & $10$ \\
Eyeglasses & $8$ \\
Everyday makeup & $5$ \\
Skin features and eyebrows & $10$ \\
Illumination & $8$ \\
Situational appearance & $3$ \\
Head coverings and hair & $2$ \\
Compound (age, appearance, eyewear, lighting) & $15$ \\
\midrule 
Total & $61$ (+ null) \\
\bottomrule
\end{tabular}
\end{table}

\paragraph{Acceptance and fallback.}
If none of the three attempts satisfies both criteria, the original crop is
retained unchanged rather than admitting a geometrically inconsistent sample, so
a failed generation never removes a training example. The null prompt has the 
same effect by design. The resulting set is therefore a mixture: approximately
$82\%$ synthesized variants, $17\%$ originals kept by fallback, and $2\%$
pass-throughs. The pipeline was applied to the $67$ training subjects; no test
subject is augmented.

\section{Decoder Architecture}

\begin{figure}[t]
  \centering
   \includegraphics[width=0.6\linewidth]{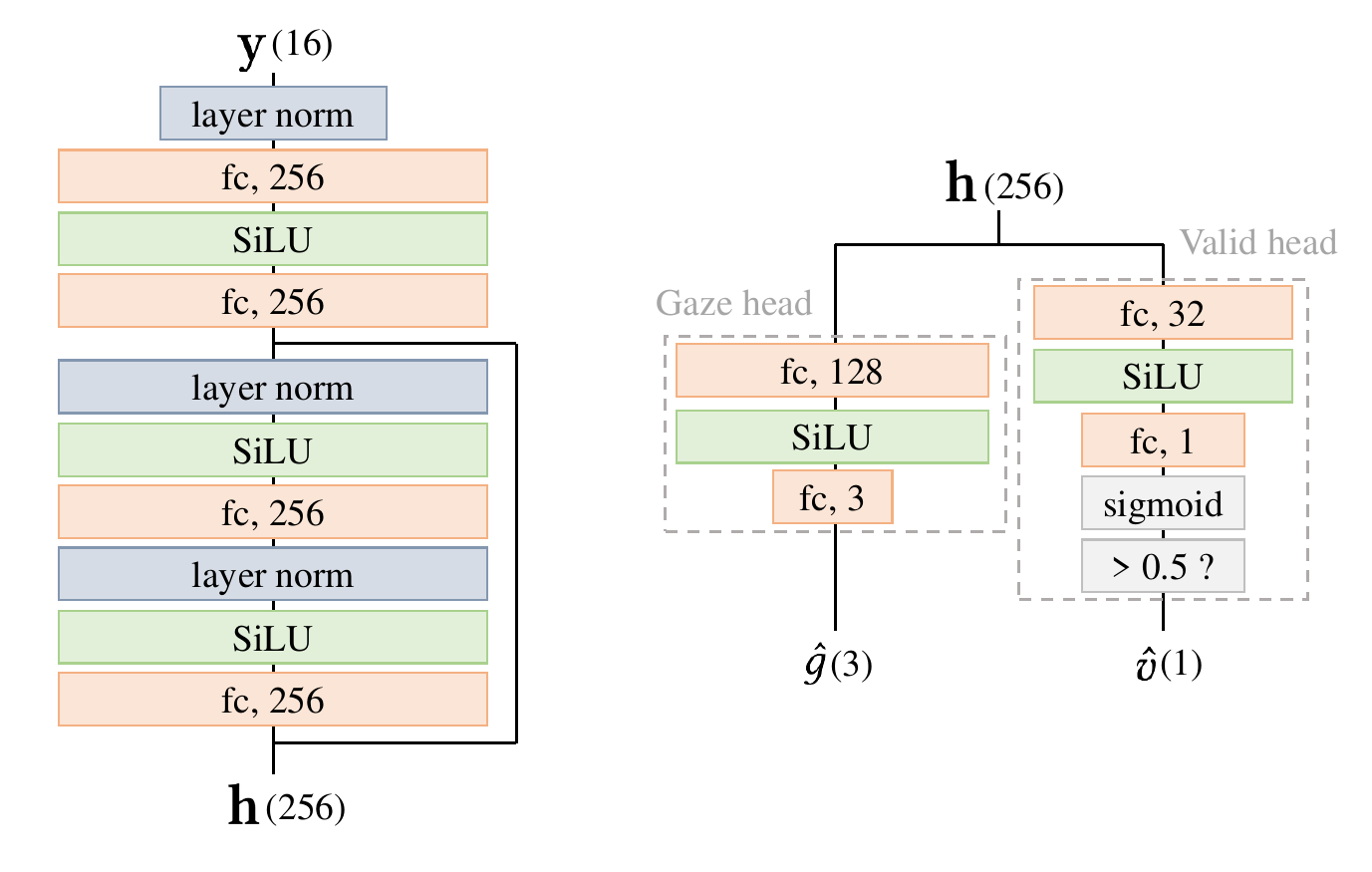}
   \caption{\textbf{Detailed schematic of the lightweight decoder $\mathcal{P}$.} (Left) The input measurement vector $\mathbf{y} \in \mathbb{R}^{16}$ is processed through a series of LayerNorm and FC-256 blocks to generate the bottleneck representation $\mathbf{h}$.
(Right) The bottleneck feature is split into a Gaze Head for 3D vector prediction and a Valid Head that utilizes a sigmoid threshold ($>0.5$) to filter artifacts such as blinks or misalignments}
   \label{fig:decoder}
\end{figure}

The digital processing component of our pipeline is designed specifically for extreme computational efficiency, enabling real-time inference on resource-constrained hardware such as mobile CPUs. The decoder $\mathcal{P}$ transforms the 16-dimensional latent measurement vector $\mathbf{y}$ into a 256-dimensional bottleneck feature $\mathbf{h}$, which is subsequently processed by two dedicated heads. The architecture of decoder is shown in Fig.~\ref{fig:decoder}.

By utilizing only $244\text{K}$ parameters and $489\text{K}$ FLOPs, the decoder achieves a forward-pass latency of only $0.22~\mathrm{ms}$ on a standard workstation CPU. This design highlights the advantage of task-driven optical sensing, as the primary feature extraction is performed passively in the optical domain, leaving only a minimal computational workload for the digital processor. 
\begin{figure}
  \centering
   \includegraphics[width=1\linewidth]{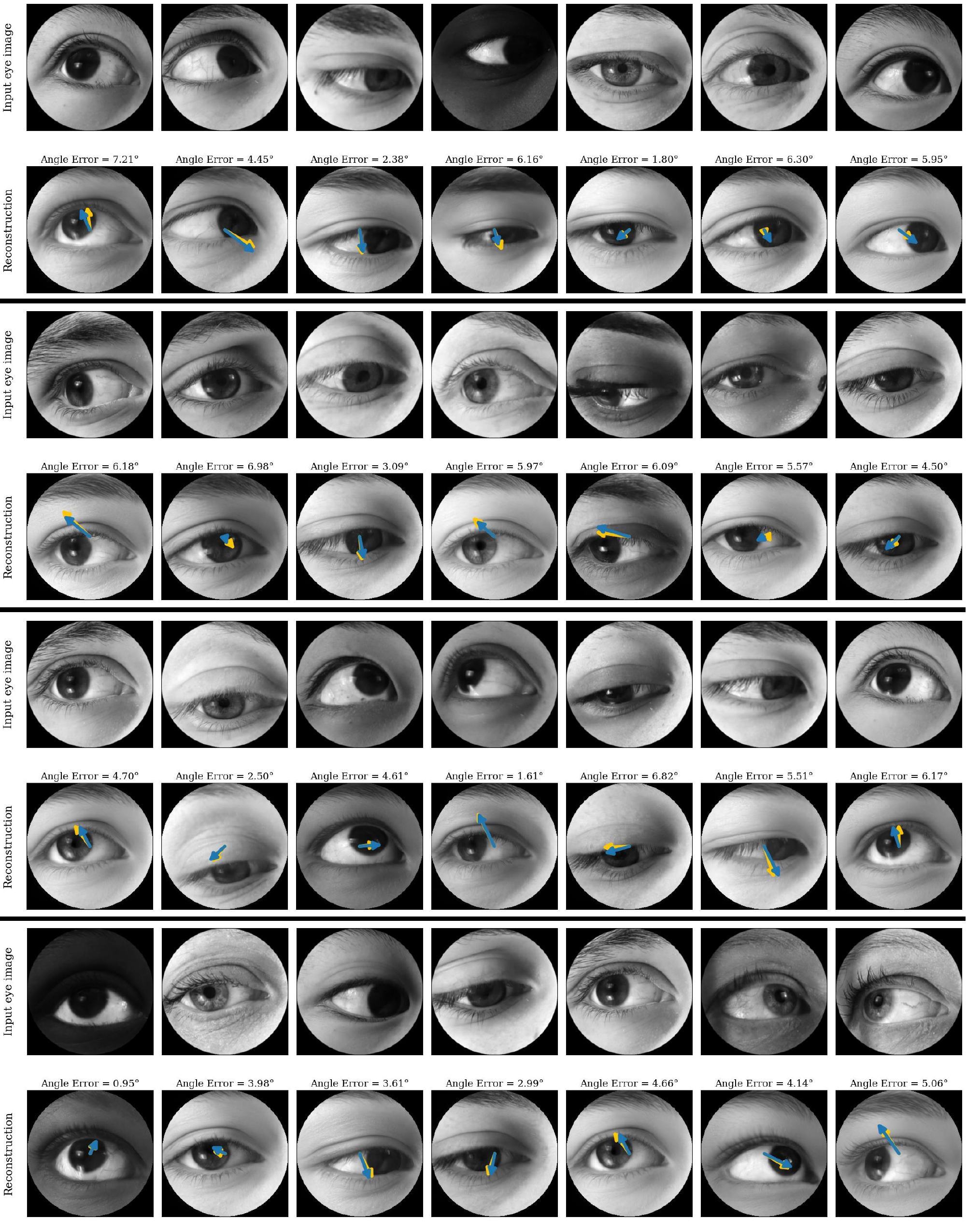}
   \caption{Gallery of latent sensing results in simulation.}
   \label{fig:sim_results}
\end{figure}
\section{Simulation Results}

To further validate the robustness of the latent-space sensing architecture, we present an expanded set of simulation results across the test set of 12 subjects. Fig.~\ref{fig:sim_results} visualizes 28 representative samples, illustrating the system's ability to generalize across varying eye shapes, skin tones, and periocular textures. An inherent advantage of our latent optical sensing paradigm is the decoupling of task-relevant information from subject identity. While the system effectively captures and retreats gaze direction with high accuracy, the resulting reconstructions in Fig.~\ref{fig:sim_results} exhibit a noticeable loss of fine-grained biometric features.

We also examined the effect of using fixed masks instead of jointly optimizing the mask with the model. With a fixed random mask, the angular error increased substantially to $11.87^\circ$. We further tested a Hadamard mask, commonly used in compressive sensing; however, this configuration led to unstable training, with the angular error remaining above $20^\circ$. These results indicate that joint mask design is critical for stable training and accurate gaze estimation in our setting.

\section{Training Details}
\label{sec:training_details}

This section gives the training configuration of the three stages introduced in
Sec.~3.1 of the main paper. All stages use grayscale $256\times256$ single-eye
crops from the ETH-XGaze training split (Sec.~3.3 main paper), with per-sample ground-truth
gaze $\gazegt$ and validity flag $\validgt$; angular errors are means over the
held-out test split.
\subsection{Generative model}
\label{sec:td_stylegan}

The generator $\generator$~\citesupp{karras2020analyzing} is trained on the
$62{,}309$ identity-augmented eye patches only (Sec.~3.4 of the main paper and
Sec.~\ref{sec:details_aug}), so that the generated images are not dominated by any specific identity. Its latent and style dimensions are both $256$, the mapping network has two layers, and the synthesis network produces $256\times256$ single-channel images from $L=14$ style vectors.

Patches are stored at $288\times288$ and augmented online in a gaze-preserving
manner: a small random translation of $\pm16$ pixels, a $256\times256$ center
crop, a circular aperture of radius $128$ (pixels outside the aperture are set to
$-1$, matching the physical aperture of the sensor), and per-sample brightness
($\pm0.2$) and contrast ($\times[0.8,1.2]$) jitter. We use Adam with learning
rate $2.5\times10^{-3}$, $\beta=(0,0.99)$, $R_1$ regularization with
$\gamma=0.8192$ and batch size $16$, converging to $\mathrm{FID}_{50k}=6.44$ at
$7{,}000$~kimg.

\subsection{Inversion teacher}
\label{sec:td_teacher}

The backbone of $\teacher$ is adapted to single-channel input in the style of
pSp~\citesupp{richardson2021encoding}: global average pooling followed by an
equalized linear layer $512\!\rightarrow\!256$ yields the compact feature
$\hteacher\in\mathbb{R}^{256}$, which is broadcast across the $L=14$ StyleGAN
layers to form the $\mathcal{W}^{+}$ code $\mathbf{w}\in\mathbb{R}^{14\times256}$
whenever synthesis is required. Two MLP heads read $\hteacher$: an angle head
($256\!\rightarrow\!128\!\rightarrow\!3$, SiLU) producing $\gaze_{T}$ and a
validity head ($256\!\rightarrow\!32\!\rightarrow\!1$) producing $\valid_{T}$.

Writing $\inimagepred = \generator(\mathbf{w} + \bar{\mathbf{w}})$ for the
reconstruction obtained by decoding the predicted latent (offset by the mean 
latent $\bar{\mathbf{w}}$) with the frozen generator, the teacher is trained with

\begin{equation}
\begin{split}
\mathcal{L}_{\teacher} =
&\lambda_{\text{ang}}
\validgt \,
\| \gaze_{T} - \gazegt \|_1
+
\lambda_{\text{cls}}
\text{BCE}(\valid_{T}, \validgt)
\\
&+
\lambda_{\text{rec}}
\validgt \,
\| \inimagepred - \inimage \|_2^2
+
\lambda_{\text{gaze}}
\validgt \,
\| E(\inimagepred) - E(\inimage) \|_1 ,
\end{split}
\end{equation}

with $\lambda_{\text{ang}} = 1$, $\lambda_{\text{cls}} = 10^{-3}$,
$\lambda_{\text{rec}} = 1$ and $\lambda_{\text{gaze}} = 0.5$, where $E$ is a
frozen single-eye gaze estimator (EfficientNet-B0~\citesupp{tan2019efficientnet},
trained separately on the same data). The last term is a perceptual \emph{gaze}
consistency: it uses no labels, but forces the reconstruction to be read by $E$
as the same direction as the real image, which keeps the inverted latent
gaze-faithful rather than merely photometrically close.

We optimize with AdamW (learning rate $10^{-4}$, weight decay $3\times10^{-3}$)
under a one-cycle schedule (peak $10^{-4}$, warm-up fraction $0.01$) for $100$k
iterations at batch size $32$ on two A100 GPUs. The encoder input is augmented
with mild blur and additive noise, while the reconstruction target and the
gaze-consistency branch always use the clean image $\inimage$. The retained
teacher attains $2.97^\circ$ on the test split.


\subsection{Learned masks and gaze decoder}
\label{sec:td_distill}

During training, each mask $\mathbf{m}_k$ of $\encodersoft$ is
parameterized as a $64\times64$ map of values in $[0,1]$ (initialized uniformly
at random), bilinearly upsampled to $256\times256$, binarized to one bit with a
straight-through estimator, and multiplied by the circular aperture. Gradients
therefore reach the continuous parameterization while the forward pass uses the
binary pattern that the optical encoder $\encoder$ can realize; the mask bank
holds $65{,}536$ trainable values.

Completing the decoder description of Sec.~3.1, the normalization layer of
$\decoder$ is a LayerNorm over $\mathbf{y}$, the upscaling block is
$16\!\rightarrow\!256\!\rightarrow\!256$ with SiLU and dropout $0.2$, and the
residual bottleneck consists of two pre-norm blocks of width $256$, yielding
$\hstudent\in\mathbb{R}^{256}$. The decoder has $244{,}324$ parameters.

In the training objective of Sec.~3.1 we set $\lambda_{\text{sup}} = 1$,
$\lambda_{\text{cls}} = 10^{-3}$, $\lambda_{\text{gaze-distill}} = 0.5$ and
$\lambda_{\text{feat-distill}} = 0.1$. The teacher always sees the \emph{clean}
image while $\encodersoft$ receives the augmented one, so the two distillation
terms additionally act as an augmentation-invariance constraint.

Optimization uses two AdamW groups --- the decoder at learning rate $10^{-4}$
with weight decay $10^{-2}$, the masks at $10^{-3}$ without weight decay --- both
on a schedule that holds the peak rate for the first $80\%$ of training and then
decays linearly to $0.1\times$, with gradient clipping at $1.0$ over the union of
both parameter sets. We train for $100$k iterations at batch size $128$ on a
single A100. Training images are augmented with random Gaussian blur (kernel
$11$, $\sigma\in[0.1,10]$) and additive Gaussian noise ($\sigma\in[0.01,0.10]$),
each applied with probability $0.5$, modeling the defocus and photon noise of the
optical encoder.

\begin{table}[t]
\centering
\small
\caption{Training configuration of the three stages.}
\label{tab:training_details}
\begin{tabular}{llll}
\toprule
& $\generator$ & $\teacher$ & $\encodersoft + \decoder$ \\
\midrule
Architecture  & StyleGAN2                 & pSp encoder $\to\mathcal{W}^{+}$ & $16$ masks + MLP \\
Optimizer     & Adam                      & AdamW                            & AdamW ($\times2$) \\
Learning rate & $2.5{\cdot}10^{-3}$       & $10^{-4}$                        & $10^{-4}$ / $10^{-3}$ \\
Weight decay  & --                        & $3{\cdot}10^{-3}$                & $10^{-2}$ / $0$ \\
Schedule      & constant                  & one-cycle                        & plateau + decay \\
Batch size    & $16$                      & $32$                             & $128$ \\
Iterations    & $8{,}000$ kimg            & $100$k                           & $100$k \\
GPUs          & $1\times$A100             & $2\times$A100                    & $1\times$A100 \\
Metric        & $\mathrm{FID}_{50k}\,6.44$ & $2.97^\circ$                    & $6.47^\circ$ \\
\bottomrule
\end{tabular}
\end{table}

\section{Per-Subject Calibration}
\label{sec:calibration}

For a budget of $K$ points we select, among the $150$ recorded samples of a
subject, the $K$ whose reference gaze best tiles the gaze range: a regular grid is
placed over the bounding box of the subject's pitch--yaw distribution and each
grid position is snapped to the nearest unused sample. The selection is
deterministic, and the remaining $150-K$ samples are held out for evaluation, so
all reported errors are measured on gaze directions never seen during 
calibration.

The decoder consumes the normalized phototransistor readings directly, its input
LayerNorm absorbing the per-sample gain and offset, so no radiometric calibration
of the sensor is required. Starting from the distilled weights, all parameters of
$\decoder$ are fine-tuned on the $K$ pairs by minimizing the mean angular error,
\begin{equation}
\mathcal{L}_{\text{cal}}
=
\frac{1}{K}\sum_{j=1}^{K}
\arccos\!\big(\cos(\gaze_j, \gazegt_j)\big),
\end{equation}
with AdamW at learning rate $10^{-4}$, batch size $16$ and gradient clipping at
$0.05$. Fitting $0.24$M parameters to as few as nine samples requires strong
regularization, so we use a weight decay of $0.5$, which together with the small
clipping threshold keeps the solution close to the distilled initialization. Each
(subject, $K$) pair is calibrated independently with a fixed seed. We run
$3{,}500$ epochs, evaluate on the held-out samples every $50$ epochs, and retain
the best checkpoint.

Table~\ref{tab:calibration} extends the budgets of the main paper to $K=18$ and
$K=21$ and reports the spread across subjects. Beyond $K=12$ the error stays
within $0.33^\circ$ of the operating point, well inside the between-subject
standard deviation of $\approx 1.3^\circ$. The residual error is therefore
dominated by between-subject variation rather than by the budget.

\begin{table}[t]
\centering
\small
\caption{Per-subject calibration of $\decodersub$ on the physical prototype: mean
angular error as a function of the calibration budget $K$, evaluated on the
$150-K$ samples held out from each subject's calibration set. Values are the mean
$\pm$ standard deviation across the $12$ subjects.}
\label{tab:calibration}
\begin{tabular}{cc}
\toprule
$K$ & MAE ($^\circ$) $\downarrow$ \\
\midrule
$9$  & $6.55 \pm 1.28$ \\
$12$ & $\mathbf{5.96 \pm 1.26}$ \\
$15$ & $5.92 \pm 1.41$ \\
$18$ & $5.69 \pm 1.29$ \\
$21$ & $5.64 \pm 1.22$ \\
\bottomrule
\end{tabular}
\end{table}

\bibliographystylesupp{unsrtnat}
\bibliographysupp{bibliography_supp}

\end{document}